\documentclass[a4paper,12pt]{article}
\usepackage[latin1]{inputenc}
\usepackage[english]{babel}
\usepackage{lineno}
\usepackage{cancel}
\usepackage[showframe=false]{geometry}
\usepackage{changepage}
\usepackage{url}

\usepackage[dvipsnames]{xcolor}
\newcommand{\brown}[1]{\textcolor{brown}{#1}}
\newcommand{\blue}[1]{\textcolor{blue}{#1}}
\newcommand{\green}[1]{\textcolor{green}{#1}}
\newcommand{\orange}[1]{\textcolor{BurntOrange}{#1}}

\renewcommand{\brown}{}
\renewcommand{\blue}{}
\renewcommand{\green}{}
\renewcommand{\orange}{}

\title{Fit without fear:  remarkable mathematical phenomena of deep learning through the prism of interpolation}
\author{Mikhail Belkin\\ Halicio\u{g}lu Data Science Institute,\\
  University of California San Diego\\ La Jolla, USA}
\date{}

\usepackage[square,sort,comma,numbers]{natbib}
\usepackage{graphicx}
\usepackage{enumitem}
\usepackage{bbm}
\usepackage{amsthm}
\usepackage{subcaption}
\usepackage{amsmath}
\usepackage{amssymb}
\usepackage{wrapfig}

\newcommand\T{{\scriptscriptstyle{\mathsf{T}}}}

\newcommand\tr{{\operatorname{tr}}}
\newcommand\vol{{\operatorname{vol}}}

\newcommand\sign{{\operatorname{sign}}}
\newcommand\E{{\mathbb{E}}}   

\renewcommand\H{{\mathcal{H}}}
\newcommand\LL{{\mathcal{L}}}
\renewcommand\S{{\mathcal{S}}}
\renewcommand\T{{\mathcal{T}}}
\newcommand\A{{\mathcal{A}}}
\newcommand\B{{\mathcal{B}}}
\newcommand\X{{\mathcal{X}}}

\newcommand\C{{\mathbb{C}}}
\newcommand\R{{\mathbb{R}}}

\def\balpha{{\boldsymbol{\alpha}}}
\def\bbeta{{\boldsymbol{\beta}}}

\def\ba{{\mathbf{a}}}
\def\bb{{\mathbf{b}}}

\def\bv{{\mathbf{v}}}
\def\bw{{\mathbf{w}}}
\def\bW{{\mathbf{W}}}
\def\bx{{\mathbf{x}}}
\def\by{{\mathbf{y}}}
\def\bz{{\mathbf{z}}}

\def\rmX{{\mathbf{X}}}

\def\ferm{f_{\rm emp}}
\def\fker{f_{\rm ker}}
\def\fkerq{f_{\rm ker,q}}
\def\fsing{f_{\rm sing}}
\def\fsimp{f_{\rm simp}}
\def\fint{f_{\rm int}}
\def\fopt{f^*}

\def\Remp{{\mathcal R}_{\rm emp}}
\def\Rtrue{{\mathcal R}}
\def\nn{{\rm 1{\text -}NN}}
\def\knn{{\rm k{\text -}NN}}
\def\lmin{\lambda_{\text min}}

\def\capH{{\mathop{cap}(\H)}}
\def\capHX{{\mathop{cap}(\H,X)}}

\theoremstyle{definition}

\begin{document}

\maketitle
{\begin{center} \footnotesize\it  In memory of Partha Niyogi, a thinker, a teacher, and a dear friend.\end{center}}

\begin{abstract}
\blue{
In the past decade the mathematical theory of machine learning has lagged far behind the triumphs of deep neural networks on practical challenges. However, the gap between theory and practice is gradually starting to close. In this paper I will attempt to assemble some pieces of the remarkable and still incomplete mathematical mosaic emerging from the efforts to understand the foundations of deep learning. 
The two key themes will be interpolation,  and its sibling, over-parameterization. Interpolation corresponds to fitting data, even noisy data, exactly. Over-parameterization enables interpolation and provides flexibility to select a right interpolating model. 
}

\blue{
As we will see, just as a physical prism separates  colors mixed within a ray of light, the figurative prism of interpolation helps to disentangle   generalization and optimization properties  within the complex picture of modern Machine Learning. This article is written with belief and hope that clearer  understanding of these issues brings us a step closer toward a general theory of deep learning and machine learning.  
}

\end{abstract}

\tableofcontents

\section{Preface} 

In recent years we have witnessed triumphs of  Machine Learning  in practical challenges from machine translation to playing chess to   protein folding. These successes rely on advances in designing and training complex neural network architectures and on availability of extensive datasets. Yet, while it is easy to be  optimistic about the potential of deep learning for  our technology and science, we may still underestimate the power of fundamental mathematical and scientific principles that can be learned from its  empirical successes. 

In what follows, I will attempt to assemble some pieces of the remarkable mathematical mosaic that is starting to emerge from the practice of deep learning.   This is an effort to capture parts of  an evolving and still elusive picture with many of the  key pieces still missing. The discussion will be largely informal, aiming to build mathematical concepts and intuitions around empirically observed phenomena. Given the fluid state of the subject and our incomplete understanding, it is necessarily a subjective, somewhat impressionistic and, to a degree, conjectural view, reflecting my understanding and  perspective. It should not be taken as a definitive description of the subject as it stands now. Instead,  it is written with the aspiration of    informing and intriguing  a mathematically minded reader and encouraging deeper and more detailed research.

\section{Introduction}

In the last decade theoretical machine learning faced a crisis. Deep learning, based on training complex neural architectures, has become  state-of-the-art for many practical problems, from computer vision to playing the game of Go to Natural Language Processing and even for basic scientific problems, such as, recently, predicting protein folding~\cite{senior2020improved}.  
\blue{Yet, the  mathematical theory of statistical learning extensively developed in the 1990's and 2000's struggled to provide a convincing explanation for its successes, let alone help in designing new algorithms or providing guidance in improving neural architectures.} This disconnect resulted in significant tensions between theory and practice. The practice of machine learning was compared to ``alchemy'', a pre-scientific pursuit, proceeding by pure practical intuition and lacking firm  foundations~\cite{alchemy}. On the other hand, a  counter-charge of practical irrelevance,  ``looking for lost keys under a lamp post, because that's where the light is''~\cite{streetlight} was leveled against the mathematical theory of learning. 

In what follows, I will start by outlining some of the reasons why classical theory failed to account for the practice of  ``modern''
machine learning. I will proceed to discuss an  emerging mathematical understanding of the observed phenomena, an understanding which points toward a reconciliation between theory and practice.

The key themes of this discussion are based on the notions of interpolation and over-parameterization, and the idea of a separation between the two regimes:
\paragraph{``Classical'' under-parameterized regimes.} The classical setting can be characterized  by limited model complexity, which does not allow arbitrary data to be fit exactly. The goal is to understand the properties of the (typically unique) classifier with the smallest loss. The standard tools include Uniform Laws of Large Numbers resulting in ``what you see is what you get'' (WYSIWYG) bounds, where the fit of classifiers on the training data is predictive of their generalization to unseen data. Non-convex optimization problems encountered in this setting typically have multiple isolated local minima, and the optimization landscape is locally convex around each minimum.  
\paragraph{``Modern'' over-parameterized regimes.}  Over-parameterized setting deals with rich model classes, where there are generically manifolds of potential {\it interpolating predictors}  that fit the data exactly. As we will discuss, some but not all of those predictors exhibit strong generalization to unseen data. Thus, the statistical question is understanding the nature of the {\it inductive bias} -- the properties that make some solutions preferable to others despite all of them fitting the training data equally well. 
In interpolating regimes, non-linear optimization problems
generically have manifolds of global minima. Optimization is  always non-convex, even locally,  yet it can often be shown to satisfy the so-called Polya\blue{k}\,-\,{\L}ojasiewicz (PL) condition guaranteeing convergence of gradient-based optimization methods.

As we will see, interpolation, the idea of fitting  the training data exactly, and its sibling over-parameterization, having sufficiently many parameters to satisfy the constraints corresponding to fitting the data, taken together provide a perspective on  some of the more surprising aspects of  neural networks and other inferential problems. 
It is interesting to point out that 
interpolating noisy data is a deeply uncomfortable and counter-intuitive concept to \brown{statistics, both theoretical and applied}, as it is traditionally concerned with over-fitting the data.  For example, in a book on non-parametric statistics \cite{gyorfi02}(page 21) the authors dismiss a certain procedure on the grounds that it ``may lead to a function which interpolates the data and hence is not a reasonable estimate''. Similarly, a popular reference~\cite{friedman2001elements}(page 194) suggests that ``a model with zero training error is overfit to the training data and will typically generalize poorly''.

\blue{Likewise, over-parameterization is alien to optimization theory, which is  traditionally more interested in convex problems  with unique solutions or non-convex problems with locally unique solutions. In contrast, as we discuss in Section~\ref{sec:conv}, over-parameterized optimization problems are in essence never convex nor have unique solutions, even locally. Instead, the  solution chosen by the algorithm depends on the specifics of the optimization process. }

To avoid confusion, it is important to emphasize  that interpolation is not {\it necessary} for good generalization. In certain models (e.g.,~\cite{hastie2019surprises}), introducing some regularization is provably preferable to fitting the data exactly. In practice, early stopping is typically used for training neural networks. It prevents the optimization process from full convergence and acts  as a type of  regularization~\cite{yao2007early}.
What is remarkable is that interpolating predictors often provide  strong generalization performance, comparable to the best possible predictors.  
Furthermore, the best practice of modern deep learning is arguably much closer to interpolation than to the classical regimes (when training and testing losses match). For example in his 2017 tutorial on deep learning\blue{~\cite{ruslantutorial}} Ruslan Salakhutdinov stated that {\it ``The best way to solve the problem from practical standpoint is you build a very big system $\ldots$ basically you want to make sure you hit the zero training error''.} While more tuning is typically needed for best performance, these ``overfitted'' systems already work well~\cite{zhang2016understanding}. 
Indeed, it appears that the largest technologically feasible networks are consistently preferable for best performance. For example, in 2016 the largest neural networks had fewer than  $10^9$ trainable parameters~\cite{canziani2016analysis}, the current (2021) state-of-the-art Switch Transformers~\cite{fedus2021switch} have over $10^{12}$ weights, over three  orders of magnitude growth in under five years!

Just as a literal physical prism separates  colors mixed within a ray of light, the figurative prism of interpolation helps to disentangle  a  blend of  properties  within the complex picture of modern Machine Learning. While significant parts are still hazy or missing  and  precise analyses are only being developed, many important pieces  are starting to fall in place.

\section{The problem of generalization}

\subsection{The setting of statistical searning}
The simplest problem of supervised machine learning is that of  classification. To construct a clich\'{e}d ``cat vs dog'' image classifier, we are given 
data $\{(\bx_i,y_i),\, \bx_i \in \X \subset \R^d, y_i \in \{-1,1\},  i=1,\ldots,n\}$, where $\bx_i$ is the vector of image pixel values and 
the corresponding  label $y_i$ is (arbitrarily) $-1$ for ``cat'', and $1$ for ``dog''. 
The goal of a learning algorithm is to construct a function $f:\R^d \to  \{-1,1\}$ that {\it generalizes} to new data, that is, accurately classifies  images unseen in training. Regression, the problem of learning general real-valued predictions, $f:\R^d \to  \R$, is formalized similarly. 

This, of course, is an ill-posed problem which needs further mathematical elucidation before a solution can be contemplated. The usual statistical assumption is that both training data and future (test) data are independent identically distributed (iid) samples from a distribution $P$ on $\R^d\times\{-1,1\}$ (defined on $\R^d\times\R$ for regression).  While the iid assumption has significant limitations, it is the simplest and most illuminating statistical setting, and we will use it exclusively. Thus, from this point of view,  the goal of Machine Learning in classification is simply to find a function, \green{known as the Bayes optimal classifier}, that minimizes the expected probability of misclassification \brown{
\begin{equation}\label{eq:bayes_opt}
  f^*\green =  \arg\min_{f:\R^d\to\R} ~\underbrace{\E_{P(\bx,y)}\, l(f(\bx),y)}_{\text{expected loss (risk)}}   
\end{equation}
Here $l(f(\bx),y)= {\bf 1}_{f(\bx)\ne y}$  is the Kronecker delta function  called $0-1$ loss  function.}  \green{The expected loss of the Bayes optimal classifier $f^*$ it called the Bayes loss \orange{or Bayes risk}.}

We note that $0-1$ \green{loss} function can be problematic due to its discontinuous nature, and is entirely unsuitable for regression, where the square loss $l(f(\bx),y)=(f(\bx)-y)^2$ is typically used. 
\green{For the square loss, the optimal predictor $f^*$ is called the regression function.}

In what follows, we will simply denote a general loss by $l(f(\bx),y)$, specifying its exact form when needed.

\subsection{The framework of empirical and structural risk Minimization}

While obtaining the optimal $\fopt$ may be the ultimate goal of machine learning, it cannot be found directly, as in any realistic setting we lack access to the underlying distribution $P$. Thus the essential question of Machine Learning is
how  $f^*$ can be  approximated {\it given the data}.
A foundational framework for addressing that question was given by V. Vapnik~\cite{Vapnik} under the name of Empirical and Structural  Risk Minimization\footnote{While \orange{empirical and structural risk optimization} are not the same, as we discuss below, both are typically referred to as ERM in the literature.}.  
The first key insight is that the data itself can serve as a proxy for the underlying distribution.  Thus, instead of minimizing the true risk $\E_{P(\bx,y)}\, l(f(\bx), y)$, we can attempt to minimize the {\it empirical risk} 
$$\brown{
\Remp(f) =\frac{1}{n}\sum_{i=1}^n l(f(\bx_i), y_i).}
$$
Even in that formulation the problem is still under-defined as infinitely many different functions minimize the empirical risk. Yet, it can be made well-posed by restricting the space of candidate functions $\H$ to make the solution unique.
Thus, we obtain the following formulation of the Empirical Risk Minimization (ERM): 
$$
\ferm = \arg\min_{f\in \H} \Remp(f) 
$$
Solving this optimization problem is called ``training''.  Of course, $\ferm$ is only useful to the degree it approximates $f^*$. While superficially the predictors $\fopt$ and $\ferm$ appear to be defined similarly, their mathematical relationship is subtle due, in particular, to the  choice of the space $\H$, the ``structural part'' of the empirical risk minimization. 

According to the discussion in~\cite{Vapnik}, ``the theory of induction'' based on the Structural Risk Minimization must meet two  mathematical requirements:
\begin{enumerate}
    \item[ULLN:] {\it The theory of induction is based on the Uniform Law of Large Numbers.}
    \item[CC:] {\it Effective methods of inference must include Capacity Control.
    }
\end{enumerate}

A uniform law of large numbers (ULLN) indicates that for any hypothesis in $\H$, the loss on the training data is predictive of the expected (future) loss:
$$
\text{ULLN:~~~} \forall f\in \H ~~~  \Rtrue(f) = \E_{P(\bx,y)}\, l(f(\bx),y) \approx R_{emp}(f) .
$$

\green{We generally expect that $\Rtrue(f) \ge \Remp(f)$, which allows ULNN to be written as a one-sided inequality, typically of the form\footnote{This is the most representative bound, rates faster and slower than $\sqrt{n}$ are also found in the literature. The exact dependence on $n$ does not change our discussion here.}
}

\begin{equation}\label{eq:ulnn}
\forall f\in \H ~~~   \underbrace{\Rtrue(f)}_{\brown{\text{expected risk}}} - \underbrace{\Remp(f)}_{\text{empirical risk}} < \underbrace{O^*\left(\sqrt{\frac{\capH}{{n}}}\right)}_{\text{capacity term}}
\end{equation}
Here $\capH$ is a measure of the {\it capacity} of the space $\H$, such as its Vapnik-Chervonenkis (VC) dimension or the covering number (see~\cite{bousquet2003introduction}), and $O^*$ can contain logarithmic terms and other terms of lower order. 
The inequality above holds with high probability over \green{the choice of} the data sample.

Eq.~\ref{eq:ulnn} is a mathematical instantiation of the  ULLN condition and directly implies 
$$
\Rtrue(\ferm) - \min_{f\in H} \Rtrue(f) <  O^*\left(\sqrt{\frac{\capH}{{n}}}\right).
$$
This guarantees that the true risk of $\ferm$ is nearly optimal for any function in $\H$, 
as long as $\capH \ll n$.

The structural condition CC is needed to ensure that $\H$ also contains functions that approximate $f^*$. Combining CC and ULLN and applying the triangle inequality, yields a guarantee that $\Remp(\ferm)$ approximates $\Rtrue(f^*)$ and the goal of generalization is achieved. 

It is important to point out that the properties ULLN and CC are in tension to each other. If the class $\H$ is too small, no $f\in \H$ will generally  be able to adequately approximate $f^*$. In contrast, if $\H$ is too large, so that $\capH$ is comparable to $n$, the capacity term is large and there is no guarantee that
$\Remp( \ferm)$ will be close to the expected risk  $\Rtrue(\ferm)$.
In that case the bound becomes tautological (such as the trivial bound that the classification risk is bounded by $1$ from above).

\begin{figure}
 \centering 
 \includegraphics[width=1\textwidth]{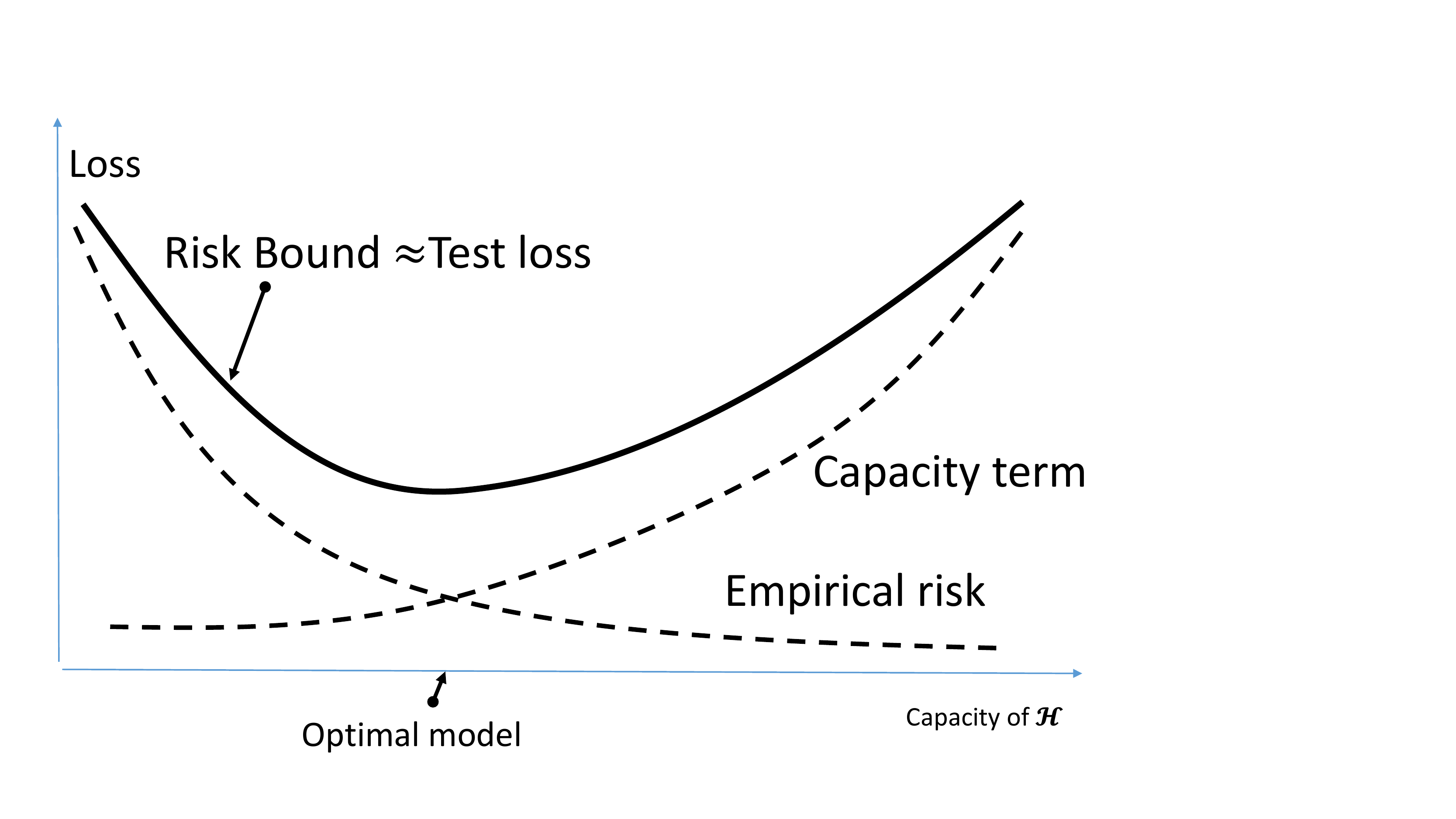}
 \caption{A classical U-shaped generalization curve. The optimal model is found by balancing the empirical risk and the capacity term. Cf.~\cite{Vapnik}, Fig.~6.2.}
 \label{fig:vapnik}
\end{figure}

Hence the prescriptive aspect of Structural Risk Minimization according to Vapnik is to enlarge $\H$ until we find the sweet spot,  a point where the empirical risk and the capacity term are balanced. This is represented by Fig.~\ref{fig:vapnik} (cf.~\cite{Vapnik}, Fig.~6.2).  

This view, closely related  to the ``bias-variance dilemma'' in statistics~\cite{geman1992neural},  had become the dominant paradigm in supervised machine learning, encouraging a rich and increasingly sophisticated line of mathematical research  uniform laws of large numbers and concentration inequalities.

\subsection{Margins theory and data-dependent explanations.}

Yet, even in the 1990's it had become clear that successes of Adaboost~\cite{freund1997decision} and neural networks were difficult to explain from the SRM or bias-variance trade-off paradigms. Leo Breiman, a prominent statistician, in his note~\cite{breiman1995reflections} from 1995 posed the question ``Why don't heavily parameterized neural networks overfit the data?''.
In particular, it was observed that increasing  complexity of classifiers (capacity of $\H$) in boosting did not necessarily lead to the expected drop of performance due to over-fitting. Why did the powerful mathematical formalism of uniform laws of large numbers fail to explain the observed evidence\footnote{This question appears as a refrain throughout the history of Machine Learning and, perhaps,  other domains.}? 

An elegant explanation known as {\it the margins theory}, 
was proposed 
in~\cite{schapire1998}. It is based on a more careful examination of the bound in Eq.~\ref{eq:ulnn}, which identifies a serious underlying issue. We observe that the bound applies to {\it any} function $f\in\H$. Yet, in the learning context,  we are not at all concerned with all functions, only with those that are plausible predictors.   Indeed, it is a priori  clear that the vast majority of predictors in standard function classes (linear functions, for example), are terrible  predictors with performance no better than chance. Whether their empirical risk matches the true risk  may be  of importance to the theory of empirical processes or to functional analysis, but is of  little concern to a ``theory of induction''. The  plausible candidate functions,  
those that are in  an appropriate sense  close to $f^*$, form a much narrower subset of $\H$. Of course, ``closeness'' needs to be carefully defined to be empirically observable without the exact prior knowledge of $f^*$. 

To give an important  special case, suppose we believe that our data are separable, so that $\Rtrue(f^*)=0$.  We can then concentrate our analysis on the subset of the hypothesis set $\H$ with small empirical loss 
$$
\H_\epsilon = \{f \in \H:~ \Remp(f) \le \epsilon\}.
$$
\brown{Indeed, since $\Rtrue(f^*)=0$, $\Remp(f^*)=0$ and hence $f^*\in \H_\epsilon$.}

The capacity  $\mathop{cap}(\H_\epsilon)$ will generally be far smaller than $\capH$ and  we thus hope for a  tighter bound.  It is important to note that the capacity $\mathop{cap}(\H_\epsilon)$ is a data-dependent quantity as $\H_\epsilon$ is defined in terms of the training data. Thus we aim to replace Eq.~\ref{eq:ulnn} with 
a  data-dependent bound:
\begin{equation}\label{eq:ulnn_data}
\forall f\in \H ~~~   \Rtrue(f) - \Remp(f) < O^*\left(\sqrt{\frac{\capHX}{{n}}}\right)
\end{equation}
\brown{where class capacity $\capHX$ depends both on the hypothesis class $\H$ and the training data $\X$. }

This important insight underlies the margins theory~\cite{schapire1998}, introduced specifically to address the apparent lack of over-fitting in boosting. 
The idea of data-dependent margin bounds has led to a line of increasingly sophisticated mathematical work on understanding data-dependent function space complexity with notions such as  Rademacher Complexity~\cite{bartlett2002rademacher}.  Yet, we note that as an explanation for the effectiveness of Adaboost, the margins theory had not been universally accepted (see, e.g.,~\cite{buja2007comment} for an interesting discussion). 

\subsection{What you see is not what you get}

It is important to note that the generalization bounds mentioned above, even the data-dependent bounds such as Eq.~\ref{eq:ulnn_data}, are  ``what you see is what you get'' (WYSIWYG): the empirical risk that you see in training approximates and bounds the true risk that you expect on unseen data, with the capacity  term providing an upper bound on the difference between  expected and \brown{empirical} risk. 

Yet, it had gradually become clear (e.g.,~\cite{neyshabur2015search}) that in modern ML, training risk and the true risk were often dramatically different and lacked any obvious connection. 
In an influential paper~\cite{zhang2016understanding} the authors demonstrate empirical evidence showing that  neural networks trained to have zero  classification risk in training do not suffer from significant over-fitting.  The authors argue that these and similar observations are incompatible with the existing learning theory and ``require rethinking generalization''. Yet, their argument does not fully rule out explanations based on data-dependent bounds such as  those in~\cite{schapire1998} which can produce nontrivial bounds for interpolating predictors if the {\it true} Bayes risk is also small.

A further empirical analysis in~\cite{belkin2018understand} made such explanations implausible, if not outright impossible.  
The experiments used a popular class of algorithms known as kernel machines, which are mathematically  predictors of the form 
\begin{equation} \label{eq:ker}
f(\bx) = \sum_{i=1}^n \alpha_i K(\bx_i,x), ~~\alpha_i \in \R
\end{equation}
Here $K(\bx,\bz)$ is a positive definite kernel function (see, e.g.,~\cite{wendland_2004} for a review), such as the commonly used  Gaussian kernel \blue{$K(\bx,\bz) = e^{-\frac{\|\bx-\bz\|^2}{2}}$}  or the Laplace kernel $K(\bx,\bz) = e^{-\|\bx-\bz\|}$. It turns out that there is a unique  predictor $\fker$ of that form  which {\it interpolates} the data:  
$$
\forall_{i=1,\ldots,n} ~~\fker (\bx_i)=y_i
$$
The coefficients $\alpha_i$ can be found analytically, by matrix inversion $\balpha= K^{-1} \by$. Here $K$ is the kernel matrix $K_{ij} = K(\bx_i,\bx_j)$, and $\by$ is the vector containing the labels $y_i$.

\begin{figure}
  \centering

  \begin{minipage}{0.45\textwidth}
    \includegraphics[width=\textwidth]{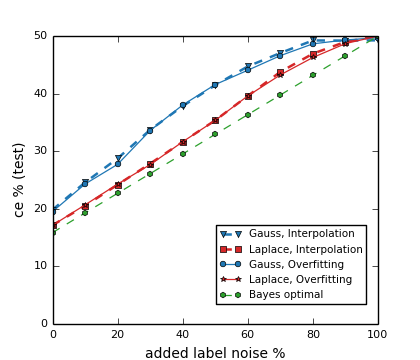}
    \subcaption{Synthetic, 2-class problem}
  \end{minipage}
  \hfill 
  \begin{minipage}[l]{0.45\textwidth}
    \includegraphics[width=\textwidth]{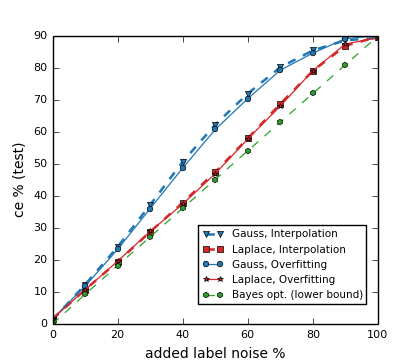}
    \subcaption{MNIST, 10-class}
    
  \end{minipage}
\caption{(From~\cite{belkin2018understand}) Interpolated  (zero training square loss), ``overfitted'' (zero training classification error), and Bayes error for datasets with added label noise. $y$ axis: test classification error. 
}
  \label{fig:interpolation_gen}
\end{figure}

Consider now a probability distribution $P$, ``corrupted'' by label noise. Specifically (for a two-class problem) with probability $q$ the label for any $\bx$ is  assigned from $\{-1,1\}$ with equal probability, and with probability  $1-q$ it is chosen according to  the original distribution $P$. Note that $P_q$ can be easily constructed synthetically  by randomizing the labels on the $q$ fraction of the training and test sets respectively.

It can be seen that the Bayes optimal classifier for the corrupted distribution $P_q$ coincides with the Bayes optimal $\fopt_P$ for the original distribution: 
$$
\fopt_{P_q} = \fopt_P.
$$
Furthermore, it is easy to check that the $0-1$ loss of the Bayes optimal predictor $\fopt_P$  computed with respect to $P_q$ (denoted by $\Rtrue_{P_q}$) is bounded from below by the noise level:
$$
\Rtrue_{P_q}(\fopt_P) \ge \frac{q}{2}
$$
It was empirically shown in~\cite{belkin2018understand} that  interpolating  kernel machines $\fkerq$ \green{(see Eq.~\ref{eq:ker})} with common Laplace and Gaussian kernels, trained to interpolate $q$-corrupted data, generalizes nearly optimally \green{(approaches the Bayes risk)} to the similarly corrupted test data. An example of that is shown in\footnote{For a ten-class problem \green{in panel (b)}, which makes the point even stronger. \green{For simplicity, we only discuss  a two-class analysis here.}} Fig.~\ref{fig:interpolation_gen}. In particular, we see that the Laplace kernel tracks the optimal Bayes error very closely,  even when as much as $80\%$ of the data  are corrupted (i.e., $q=0.8$).

Why is it surprising from the WYISWYG bound point of view? For simplicity, suppose $P$ is deterministic ($\Rtrue(\orange{\fopt_P})=0$), which is essentially  the case \green{[FOOTNOTE MOVED]} in  Fig.~\ref{fig:interpolation_gen}, Panel (b).   In that case \green{(for a two-class problem)},  $\Rtrue_{P_q}(\orange{\fopt_P}) = \frac{q}{2}$.  

$$\Rtrue_{P_q}(\fkerq)\, \green{\ge}\, \Rtrue_{P_q}(\orange{\fopt_P})  = \frac{q}{2}.
$$ 
\green{On the other hand}
$\Remp(\fkerq) =0$ and \green{hence for the left-hand side in Eq.~\ref{eq:ulnn_data} we have}
$$
\green{\Rtrue_{P_q}(\fkerq)}- \orange{\underbrace{\Remp(\green{\fkerq})}_{=0}} \green{=\Rtrue_{P_q}(\fkerq)} \,\green{\ge}\, \frac{q}{2}
$$

\green{To explain good empirical performance of $\fkerq$, a bound like Eq.~\ref{eq:ulnn_data} needs to be both } {\it correct} and {\it nontrivial}. \green{Since the left hand side is at least $\frac{q}{2}$ and observing that $\Rtrue_{P_q}(\fkerq)$ is upper bounded by the loss of a random guess, which is $1/2$ for a two-class problem,} we must have 
\begin{equation}\label{eq:}
 \green{\frac{q}{2} \underbrace{\le}_{\text{correct}} O^*\left(\sqrt{\frac{\capHX}{{n}}}\right) \underbrace{\le}_{\text{nontrivial}} \frac{1}{2}}
\end{equation}

Note that  such a bound would require the multiplicative coefficient in $O^*$ to be tight  within a multiplicative factor $1/q$ (which is $1.25$  for $q=0.8$). No such general bounds are known.  In fact,  typical bounds include logarithmic factors and other multipliers making really tight estimates impossible. More conceptually, 
it is hard to see how such a bound {\it can} exist, as  the capacity term would need to ``magically'' know\footnote{\brown{This applies to the usual capacity definitions based on norms, covering numbers and similar mathematical objects. In principle, it may be possible to  ``cheat'' by letting capacity depend on complex manipulations with the data, e.g., cross-validation. \green{This requires a different type of analysis (see~\cite{negrea2020indefense,zhou2020onuniform} for some recent attempts)
and raises the question of what may be considered a  useful generalization bound. We leave that discussion for another time. }
}}
about the level of noise $q$ in the probability distribution. 
Indeed, a strict mathematical proof of incompatibility of generalization with uniform bounds was recently given in~\cite{nagarajan2019uniform} under certain specific settings. The consequent work~\cite{bartlett2020failures} proved that no good bounds can exist for a broad range of models.

Thus we see that strong generalization performance of classifiers that interpolate  noisy data is incompatible with 
 WYSIWYG bounds, independently of the nature of the capacity term.

\subsection{Giving up on WYSIWYG, keeping theoretical guarantees}

So  can we provide statistical guarantees for classifiers that interpolate noisy data? 
 
Until very recently there had not been many. 
\blue{In fact, the only common 
  interpolating algorithm with statistical guarantees for noisy data is the well-known 1-NN rule\footnote{In the last two or three years there has been significant progress on interpolating guarantees for classical algorithms like linear regression and kernel methods (see the discussion and references \brown{below}). However, traditionally analyses nearly always used regularization which precludes interpolation.}. 
  Below we will go over a sequence of three progressively more statistically powerful nearest neighbor-like interpolating predictors, starting with the classical 1-NN rule, and going to simplicial interpolation and then to general weighted nearest neighbor/Nadaraya-Watson schemes with singular kernels.
}

\subsubsection{\blue{The peculiar case of 1-NN}}

 Given an input $\bx$, $\nn(\bx)$ outputs the label for the closest (in Euclidean or another appropriate distance)  training example.  

While the 1-NN rule is among the simplest and most classical prediction rules both for classification and regression, it has several striking aspects which are not usually emphasized in standard treatments:
\begin{itemize}
    \item It is an interpolating classifier, i.e., $\Remp(\nn) = 0$.
    \item Despite ``over-fitting'', 
     classical analysis in~\cite{cover1967nearest}  shows that the classification risk  of ${\cal R}(\nn)$  is (asymptotically as $n \to \infty$)  bounded from above by $2\blue{\cdot}\Rtrue(f^*)$, where $f^*$ is the Bayes optimal classifier defined by Eq.~\ref{eq:bayes_opt}. 
    
    \item Not surprisingly, given that it is an interpolating classifier, there  no ERM-style analysis of 1-NN. 
\end{itemize}

It seems plausible that the remarkable interpolating nature of 1-NN had been written off by the statistical learning community as an aberration due to its high excess risk\footnote{Recall that the excess risk of a classifier $f$ is the difference between the risk of the classifier and the risk of the optimal predictor $\Rtrue(f) - \Rtrue(\fopt)$.}. 
 As we have seen, the risk of 1-NN can be a factor of two worse than the risk of the optimal classifier. 
The standard prescription for improving performance  is to use k-NN, an average of $k$ nearest neighbors, which  no longer interpolates. As $k$ increases (assuming $n$ is large enough), the excess risk decreases as does the difference between the empirical and expected risks. Thus, for large $k$ \blue{(but still much smaller than $n$)} we have, seemingly in line with the standard ERM-type bounds, 
$$
\Remp(\knn) \approx \Rtrue(\knn)\approx\Rtrue(f^*).$$

 It is perhaps ironic that an outlier feature of $1$-NN rule, shared with no other common methods in the classical \brown{statistics} literature (except for the 
 relatively unknown work~\cite{devroye1998hilbert}),
 \brown{may be one of the cues}  to understanding modern deep learning.

\subsubsection{\blue{Geometry of simplicial interpolation and the blessing of dimensionality} }

Yet, a  modification of 1-NN different from k-NN maintains its interpolating property while achieving near-optimal excess risk, at least in when the dimension is high. The algorithm is simplicial interpolation~\cite{halton1991simplicial}
 analyzed statistically in~\cite{belkin2018overfitting}. 
Consider a triangulation of the data, \brown{$\bx_1,\ldots,\bx_n$,} that is a partition of the convex hull of the data into a set of $d$-dimensional simplices so  that:
\begin{enumerate}
    \item Vertices of each simplex are data points.
    \item For any data point $\bx_i$ and simplex $s$, $\bx_i$ is either a vertex of $s$ or does not belong to $s$.
\end{enumerate}

The exact choice of the triangulation turns out to be  unimportant as long as the size of each simplex is small enough. This is guaranteed by, for example, the well-known Delaunay triangulation.

Given a multi-dimensional triangulation, we define $\fsimp(x)$, the simplicial interpolant, to be a function which is linear within each simplex  and such that $\fsimp(x_i)=y_i$. It is not hard to check that $\fsimp$ exists and is unique.

It is worth noting that in one dimension simplicial interpolation based on the Delaunay triangulation is equivalent to 1-NN for classification. 
Yet,  when the dimension $d$ is  high enough, simplicial interpolation is nearly optimal both for classification and regression. Specifically, it is was shown in Theorem 3.4 in~\cite{belkin2018overfitting} (Theorem 3.4) that simplicial interpolation benefits from a {\it blessing of dimensionality}. For large $d$, the excess risk of $\fsimp$ decreases with dimension:
$$\Rtrue(\fsimp) - \Rtrue(f^*) = O\left(\frac{1}{\sqrt{d}} \right).
$$ 
Analogous results hold for regression, where the excess risk is similarly the difference between \orange{the loss of} a predictor and  \orange{the loss of the} (optimal) regression function.
Furthermore,  for classification, under additional conditions $\sqrt{d}$ can be replaced by $e^{d}$ in the denominator. 

Why does this happen? How can an interpolating function be nearly optimal despite the fact that it fits noisy data and why does increasing dimension help? 

The key observation is that incorrect predictions are localized in the neighborhood of ``noisy'' points, i.e.,\,those points where $y_i=\fsimp(\bx_i)\ne f^*(\bx_i)$. 
To develop an intuition, consider the following simple example. Suppose   that $x_1,\ldots,x_{d+1} \in \R^d$ are vertices of a standard $d$-dimensional simplex $s_d$: $$\bx_i=(0,\ldots,\underbrace{1}_{i},\ldots,0),~ i=1,\ldots,d, ~~\blue{\bx}_{d+1}=(0,\ldots,0)
$$   
Suppose \green{also that} the probability distribution is uniform on the simplex (the convex hull of $\bx_1,\ldots\bx_{d+1}$) and the ``correct'' labels are identically $1$. As our training data, we are given $(\bx_i,y_i)$, where $y_i=1$, except for 
 the one vertex, which is ``corrupted by noise'', so that $y_{d+1}=-1$. 
 It is easy to verify that
 $$
 \fsimp(\bx) = \sign\, (2\sum_{i=1}^d (\bx)_i -1).
 $$
 We see that $\fsimp$ coincides with $f^*\equiv 1$ in the simplex except for the set  $s_{1/2} = \{\bx:\sum_{i=1}^d\bx_i \le \blue{1/2}\}$, which is equal to the simplex  $\frac{1}{2}s_d$ and thus 
 $$
 \vol(s_{1/2}) = \frac{1}{2^d}\vol(s_d)
 $$ 
 
 \begin{wrapfigure}{r}{0.4\textwidth}
  \vskip-1.5cm
    \includegraphics[width=.6\textwidth]{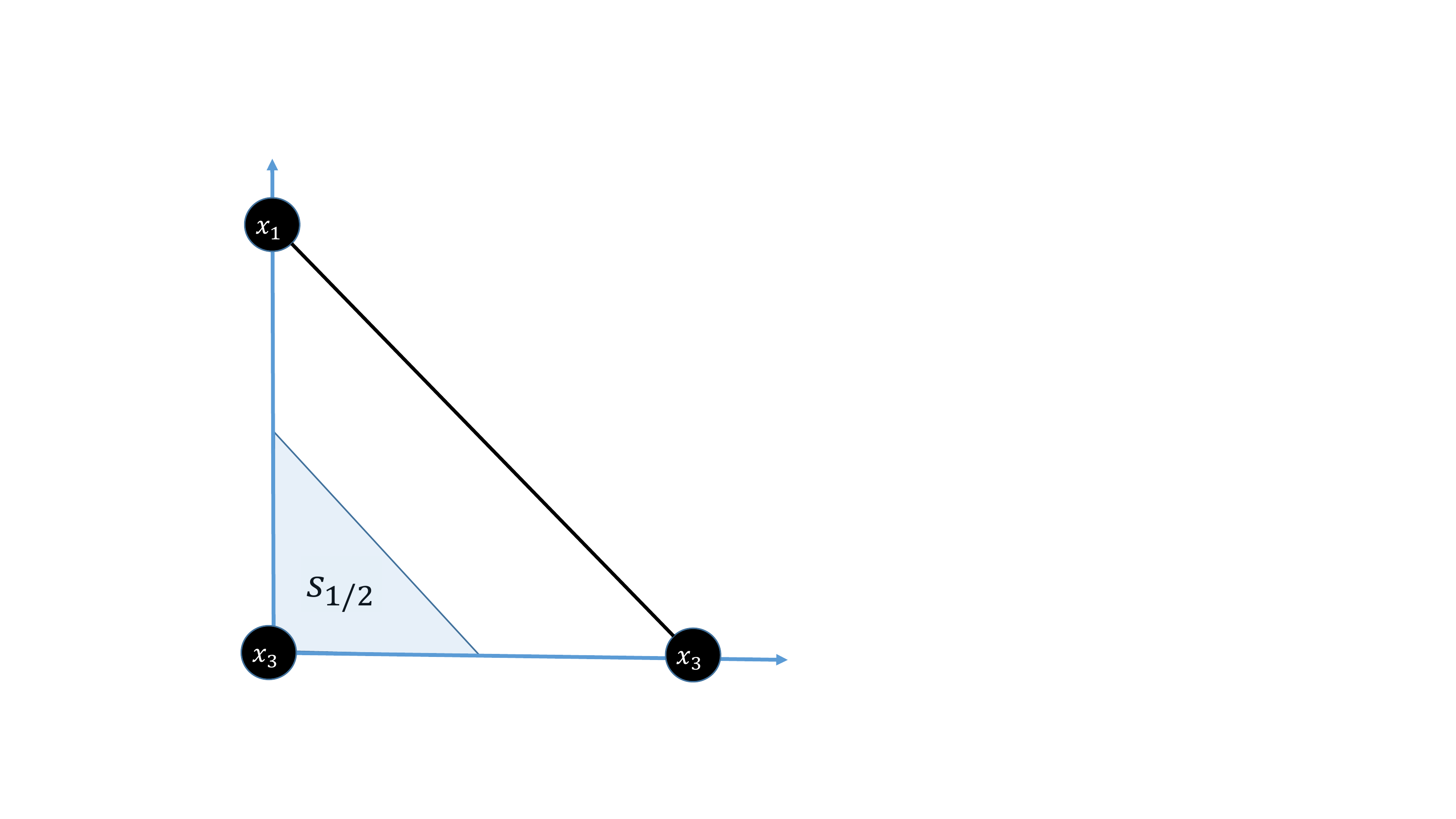}
 \caption{\green{The set of points $s_{1/2}$  where $\fsimp$ deviates from the optimal predictor $f^*$.}}\label{fig:simplex}
\end{wrapfigure}

We see that  the interpolating predictor $\fsimp$ is different from the optimal, \green{but} the difference is highly  localized around the ``noisy'' vertex, while at most points \green{within $s_d$} their predictions coincide. This is illustrated geometrically in Fig.~\ref{fig:simplex}.
The reasons for the blessing of dimensionality also become clear, as small neighborhoods in high dimension have smaller volume relative to the total space. Thus, there is more freedom and flexibility for the noisy points to be localized. 

\begin{figure}
  \begin{center}
    \includegraphics[width=\textwidth]{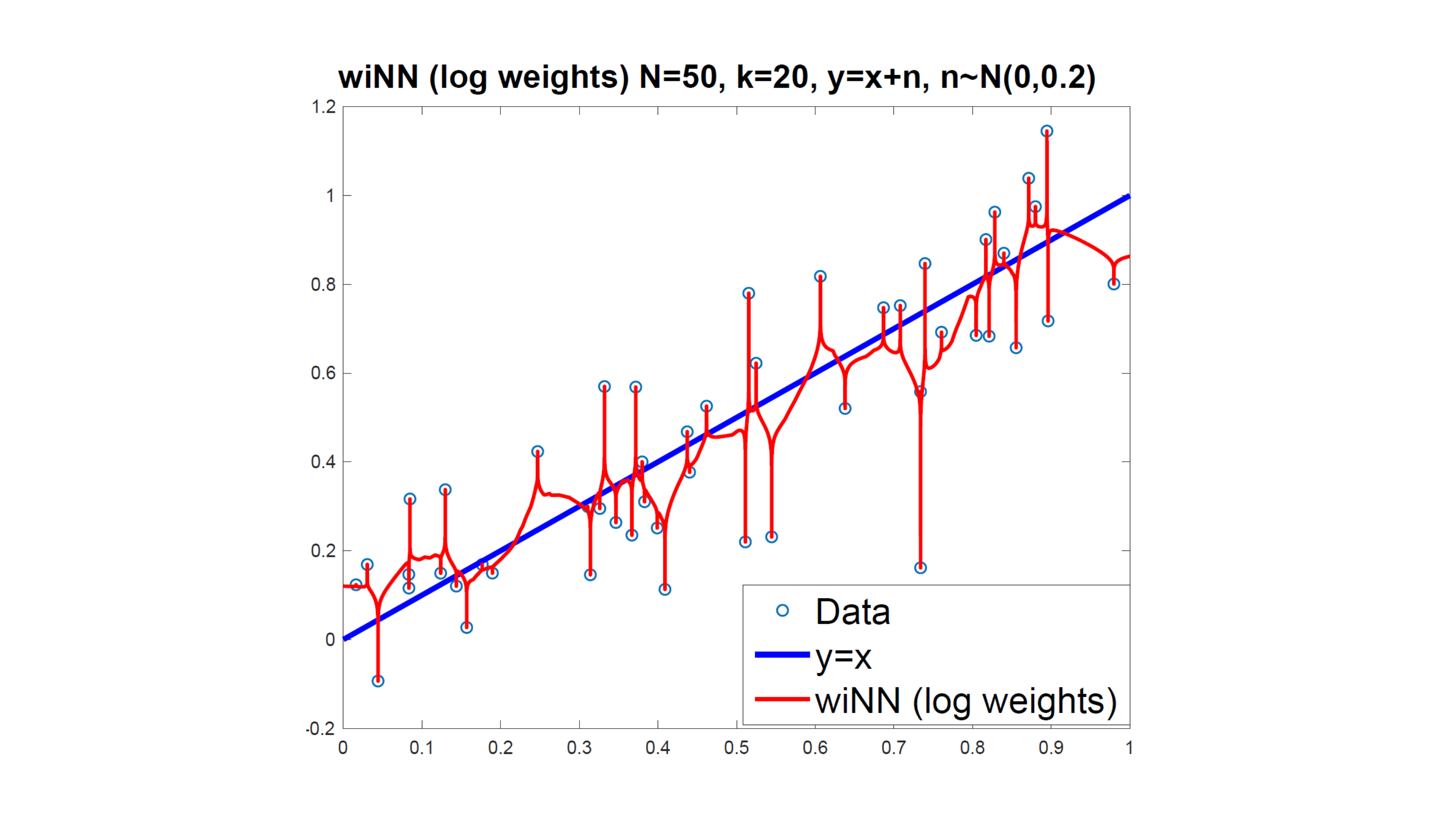}
  \end{center}
 \caption{Singular kernel for regression. \green{Weighted and interpolated nearest neighbor (wiNN) scheme. \green{Figure credit: Partha Mitra.} }}\label{fig:wiNN}
\end{figure}

\subsubsection{\blue{Optimality of k-NN with singular weighting schemes}}\label{sec:singular_ker}
 
 While simplicial interpolation improves on $\nn$ in terms of the excess loss, it is still not {\it consistent}. In high dimension $\fsimp$  is near $f^*$ but does not converge to $f^*$ as  $n\to\infty$. Traditionally, consistency and rates of convergence have been a central object of statistical investigation. The first result in this direction is~\cite{devroye1998hilbert}, which showed statistical consistency of a certain kernel regression scheme, closely related to Shepard's inverse distance interpolation~\cite{shepard1968two}.
 
 It turns out that a similar interpolation scheme based on weighted $k$-NN can be shown to be consistent for both regression and classification and indeed to be optimal in a certain statistical sense (see~\cite{belkin2018overfitting} for convergence rates for regression and classification and the follow-up work~\cite{belkin2018does} for optimal rates for regression). The scheme can be viewed as a type of Nadaraya-Watson~\citep{nadaraya1964estimating,watson1964smooth} predictor. It can be described as follows.  Let
 $K(\bx,\bz)$ be a singular kernel, such as
 $$
 K(\bx,\bz)=\frac{1}{\|\bx-\bz\|^{\alpha}},~\alpha>0, 
 $$
 with an appropriate choice of $\alpha$.
 Consider the weighted nearest neighbor predictor
\begin{align*}
  \fsing(\bx) & = \frac{\sum_{i=1}^k K(\bx,\bx_{(i)}) y_{(i)}}{\sum_{i=1}^k K(\bx,\bx_{(i)})} .
\end{align*}
Here the sum is taken over the $k$ nearest neighbors of $\bx$, $\bx_{(1)},\ldots,\bx_{(k)}$. While the kernel $K(\bx,\bx_{(i)})$ is infinite \green{at $x=\bx_i$}, it is not hard to see that $\fsing(x)$ involves a ratio that can be defined everywhere due to the cancellations between the singularities in the numerator and the denominator. It is, furthermore,  a continuous function of $\bx$. 
Note that for classification it suffices to simply take the sign of the numerator  $\sum_{i=1}^k K(\bx,\bx_{(i)}) y_{(i)}$ as the denominator is positive.

To better understand how such an unusual scheme can be consistent for regression, consider an example 
 shown in Fig.~\ref{fig:wiNN} for one-dimensional data sampled from a noisy linear model: $y =x +\epsilon$, where $\epsilon$ is normally distributed noise. Since  the predictor $\fsing(x)$  fits the noisy data exactly, it is far from optimal on the majority of data points. Yet, the prediction is close to optimal for most points in the interval $[0,1]$! In general, as $n\to \infty$, the fraction of those points tends to $1$.
 
 We will discuss this phenomenon further in connection to adversarial examples in deep learning in Section~\ref{sec:adversarial}.

\subsection{Inductive biases and the Occam's razor}

The realization that, contrary to deeply ingrained statistical intuitions, fitting noisy training data exactly  does not necessarily result in poor generalization, inevitably leads to quest for a new framework for a ``theory of induction'', a paradigm not reliant  on uniform laws of large numbers and not requiring empirical risk to approximate the true risk. 

While, as we  have seen, interpolating classifiers can be statistically  near-optimal or optimal, the predictors discussed above appear to be different from those widely used in ML practice.  Simplicial interpolation, weighted nearest neighbor or Nadaraya-Watson  schemes  do not require training and can be termed {\it direct} methods. In contrast, common practical algorithms from linear regression to kernel machines to neural network\green{s} are ``inverse methods'' based on optimization. 
 These algorithms typically rely   on {\it algorithmic empirical risk minimization}, where a loss function $\Remp(f_\bw)$ is minimized via a specific algorithm, such as stochastic gradient descent (SGD) on the weight vector $\bw$. 
Note that there is a crucial and sometimes overlooked difference between the empirical risk minimization as an {\it algorithmic process} and  the Vapnik's ERM paradigm for generalization, which is algorithm-independent. 
This distinction becomes important in over-parameterized regimes, where the hypothesis space $\H$ is rich enough to fit any data set\footnote{Assuming that $\bx_i \ne \bx_j$, when $i\ne j$.}  of cardinality $n$. The key insight is to separate ``classical'' {\it under-parameterized regimes} where there is typically no $f\in \H$, such that $\Rtrue(f)=0$ and ``modern'' over-parameterized settings where there is a (typically large)  set $\S$  of predictors that interpolate the training data 
\blue{
\begin{equation}\label{eq:interp_subset}
\S= \{f\in \H:\Rtrue(f)=0\}.
\end{equation}}
First observe that an interpolating learning algorithm
$\A$ selects  a specific predictor $f_\A\in \S$. Thus we are faced with the issue of the {\it inductive bias}: why do solutions, such as those obtained by neural networks and kernel machines, generalize, while other possible solutions do not\footnote{The existence of non-generalizing solutions is immediately clear by considering  over-parameterized linear predictors. Many linear functions fit the data -- most of them generalize poorly.}.
Notice that this question cannot be answered through the training data alone, as any $f \in \S$ fits  data equally well\footnote{We note that inductive biases are present in any inverse problem. Interpolation simply  isolates this issue.}.
While no conclusive recipe for selecting the optimal $f \in \S$ yet exists, it can be posited that an appropriate notion of functional smoothness plays a key  role in that choice. As argued in~\cite{belkin2019reconciling}, the idea of maximizing functional smoothness subject to interpolating the data  represents a very pure form of the Occam's razor (cf.~\cite{Vapnik,blumer1987occam}). Usually stated as
\begin{center}{\it Entities should not be multiplied beyond necessity,}
\end{center} 
the Occam's razor implies that the simplest explanation consistent with the evidence should be preferred. In this case fitting the data corresponds to consistency with evidence, while the smoothest function is ``simplest''.
To summarize, the ``maximum \brown{smoothness}'' guiding principle can be formulated as: 
\begin{center}{\it Select the smoothest function, according to some notion of functional smoothness,  among those that  fit the data perfectly.} 
\end{center}
We note that kernel machines described above (see Eq.~\ref{eq:ker}) fit this paradigm precisely. Indeed, for every positive definite kernel function $K(\bx,\bz)$, there exists a Reproducing Kernel Hilbert Space ( functional spaces, closely related  to  Sobolev spaces, see~\cite{wendland_2004}) $\H_K$, with norm $\|\cdot\|_{\H_K}$ such that 
\begin{equation}\label{eq:kernel}
\fker(\bx) = \arg\min_{\forall_i f(\bx_i)=y_i} \|f\|_{\H_K}
\end{equation}

We proceed to discuss how this idea may apply to training more complex variably parameterized models including neural networks. 

\subsection{The Double Descent phenomenon}

A hint toward a possible theory of induction is provided by the 
\begin{figure}
 \centering 
 \includegraphics[scale=1.7]{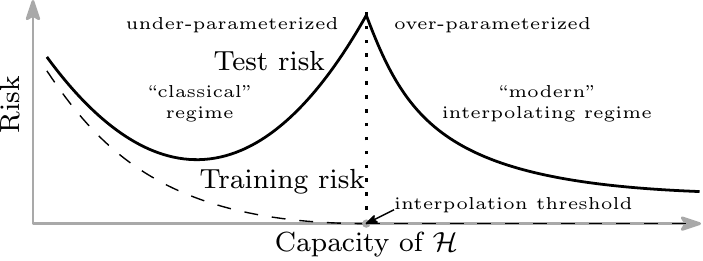}
 \caption{Double descent generalization curve (figure from~\cite{belkin2019reconciling}). Modern and classical regimes are separated by the interpolation threshold.}\label{fig:double_descent}
\end{figure}
double descent generalization curve (shown in Fig.~\ref{fig:double_descent}), a pattern proposed in~\cite{belkin2019reconciling} as a replacement for the classical U-shaped generalization curve  (Fig.~\ref{fig:vapnik}). 

When the capacity of a hypothesis class $\H$ is below the {\it interpolation threshold}, not enough to fit arbitrary data,
learned predictors follow the classical U-curve from Figure~\ref{fig:vapnik}.
The shape of the generalization curve undergoes a qualitative change when the capacity of $\H$ passes the { interpolation threshold}, i.e., becomes large enough to interpolate the data.
Although predictors at the interpolation threshold typically have high risk, further increasing the number of parameters (capacity of $\H$)  leads to improved generalization.
The double descent  pattern has been empirically demonstrated for a broad range of datasets and algorithms, including  modern deep neural networks~\cite{belkin2019reconciling,spigler2018jamming,nakkiran2019deep} and observed earlier for linear models~\cite{loog2020brief}. The ``modern'' regime of the curve, the phenomenon that large number of parameters often do not lead to over-fitting has historically been observed in boosting~\cite{schapire1998,wyner2017explaining} and random forests, including interpolating random forests~\cite{cutler2001pert} as well as in neural networks~\cite{breiman1995reflections, neyshabur2015search}.

Why should predictors from richer classes perform better given that they all fit data equally well? Considering an inductive bias based on smoothness provides an explanation for this seemingly counter-intuitive phenomenon as larger spaces contain will generally contain ``better'' functions.
Indeed, consider a hypothesis space $\H_1$ and a larger space $\H_2, \H_1 \subset\H_2$. The corresponding subspaces  of interpolating predictors, $\S_1 \subset \H_1$ and $\S_2 \subset H_2$, are also related by inclusion: $\S_1 \subset \S_2$. Thus, if  $\|\cdot\|_s$ is a functional norm, or more generally, any functional, we see that 
$$
\min_{f \in S_2} \|f\|_s  \le \min_{f \in S_1} \|f\|_s
$$
Assuming that $\|\cdot\|_s$ is the ``right'' inductive bias, measuring smoothness (e.g., a Sobolev norm), we expect the minimum norm predictor from $\H_2$, $f_{\H_2}=\arg\min_{f \in S_2} \|f\|_s$ to be superior  to that from \green{$\H_1$}, $f_{\H_1}=\arg\min_{f \in S_1} \|f\|_s$. 

\begin{figure}
 \centering 
 \includegraphics[scale=.45]{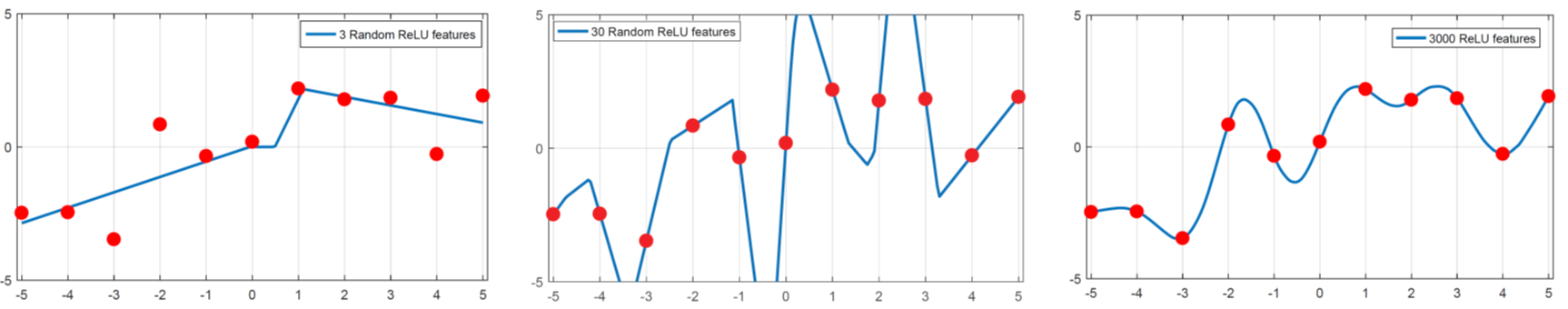}
 \caption{\brown{Illustration of double descent for Random ReLU networks in one dimension. Left: Classical under-parameterized regime ($3$ parameters). Middle: Standard over-fitting, slightly above the interpolation threshold ($30$ parameters). Right: ``Modern'' heavily over-parameterized regime ($3000$ parameters). }}\label{fig:ReLU_double_descent}
\end{figure}

\brown{A visual illustration for double descent and its connection to smoothness is provided in  Fig.~\ref{fig:ReLU_double_descent} within the random ReLU  family of models in one dimension. A very similar Random Fourier Feature family is described in more mathematical detail below.\footnote{\brown{The Random ReLU family consists of piecewise linear functions of the form $f(\bw,x)=\sum_k w_k\min(v_kx+b_k,0)$ where $v_k,b_k$ are fixed random values. While it is quite similar to RFF, it produces better visualizations in one dimension.}  
} The left panel shows  what may be considered a good fit for a model with a small number of parameters. The middle panel, with the number of parameters slightly larger than the minimum necessary to fit the data, shows textbook over-fitting. However increasing the number of parameters further results in a  far more reasonably looking curve.
While this curve is still piece-wise linear due to the nature of the model, it appears completely smooth. Increasing the number of parameters to infinity will indeed yield a differentiable function (a type of spline), although the difference between $3000$ and infinitely many parameters is not visually perceptible.  
As discussed above, over-fitting appears in a range of models around the interpolation threshold which are complex but yet not complex enough to allow smooth structure to emerge. Furthermore, low complexity parametric models and non-parametric (as the number of parameters approaches infinity) models  coexist within the same family on different sides of the interpolation threshold.
}

\paragraph{Random Fourier features.} \label{sec:rff}

Perhaps the  simplest \brown{mathematically} and most illuminating example  of the double descent phenomenon is based on Random Fourier Features 
(\emph{RFF})~\cite{rahimi2008random}.
The RFF model family $\H_m$ with $m$ (complex-valued) parameters  consists of functions \blue{$f \colon \R^d \to \C$ of the form
\[
  f(\bw,\bx) = \sum_{k=1}^{m} w_k e^{\sqrt{-1} \langle \bv_k,\bx \rangle}
\]}
where the vectors $\bv_1,\dotsc,\bv_m$ are fixed weights with values sampled independently from the standard normal distribution on $\R^d$. \blue{The vector $\bw = (w_1,\ldots,w_m) \in \C^m \cong\R^{2m}$ }consists of  trainable parameters.
$f(\bw,\bx)$ can be viewed as a neural network with one hidden layer of size $m$ and fixed first layer weights \blue{(see Eq.~\ref{eq:generalnn} below for a general definition of a neural network)}.

Given data $\{\bx_i,y_i\},i=1,\ldots,n$, we can fit $f_m \in \H_m$ by linear regression on the coefficients $\bw$.  In the overparameterized regime linear regression is given by minimizing the norm under the interpolation constraints\footnote{As opposed to the under-parameterized setting when linear regression it simply minimizes the empirical loss over the class of linear predictors.}: 
$$
f_m(\bx) = \arg\min_{f\in\H_m,\, f(\bw,\bx_i)=y_i} \|w\| .
$$

It is shown in~\cite{rahimi2008random} that 
$$
\lim_{m\to \infty} f_m(\bx) =  \arg\min_{f \in \S \subset\H_K} \|f\|_{\H_K} =: \fker(\bx) 
$$
Here $\H_K$ is the Reproducing Kernel Hilbert Space corresponding to the Gaussian kernel $K(\bx,\bz)=\exp(-\|\bx-\bz\|^2)$ and $\S \subset \H_K$ is the manifold of interpolating functions in $\H_K$.
Note that $\fker(\bx)$ defined here is the same function defined in $Eq.~\ref{eq:kernel}$. This equality is known as the Representer \green{Theorem~\cite{kimeldorf1970correspondence,wendland_2004}}.  

\begin{figure}
 \centering 
 \includegraphics[width=1\textwidth]{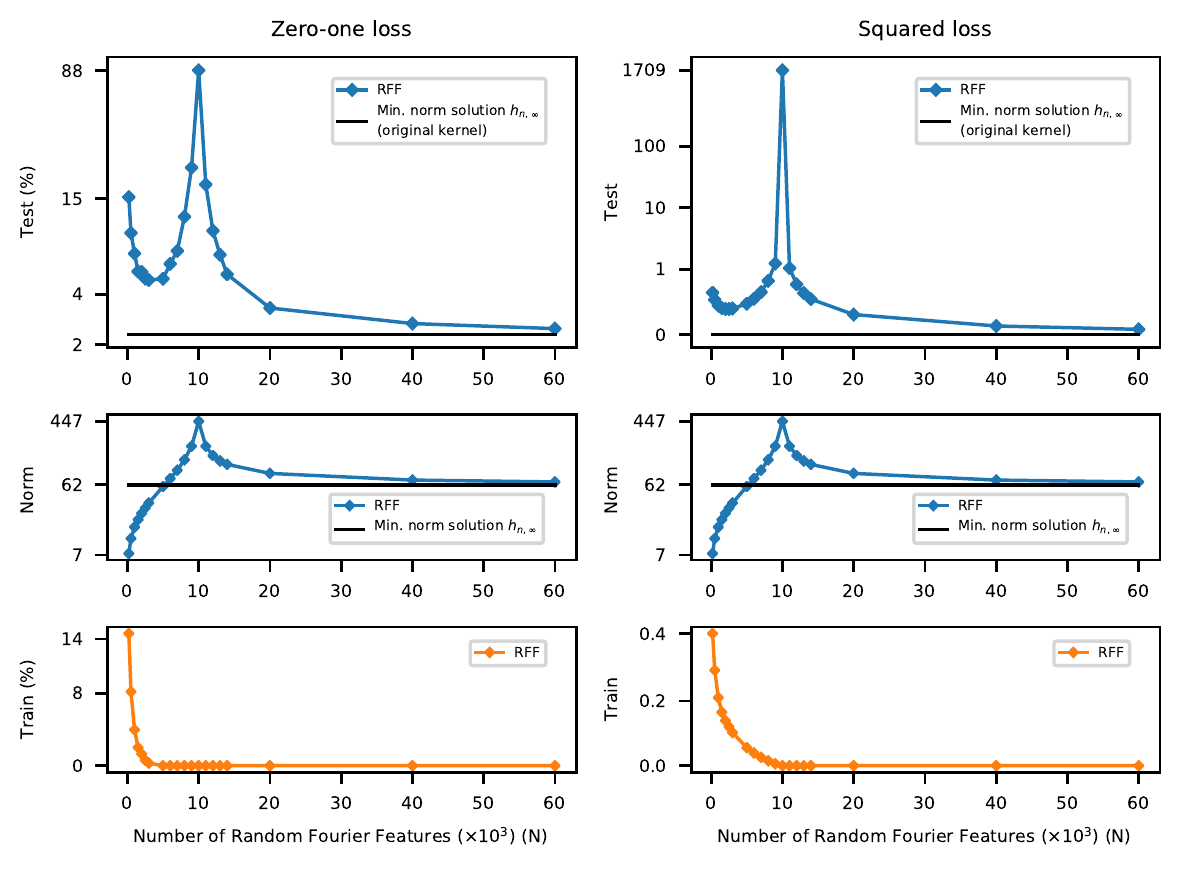}
 \caption{Double descent generalization curves and norms  for Random Fourier Features on a subset of MNIST (a 10-class hand-written digit image  dataset). Figure from~\cite{belkin2019reconciling}.}
 \label{fig:rff_norm}
\end{figure}

We see that  increasing the number of parameters $m$ expands the space of interpolating classifiers in $\H_m$ and allows to obtain progressively better approximations of the ultimate functional smoothness minimizer $\fker$. Thus adding parameters in the over-parameterized setting leads to solutions with smaller norm, in contrast to under-parameterized classical world when more parameters imply norm increase. The norm of the weight vector $\|\bw\|$ asymptotes to the true functional norm of the solution $\fker$ as $m \to \infty$. 
This is verified experimentally in Fig.~\ref{fig:rff_norm}. We see that the generalization curves for both 0-1 loss and the square loss follow the double descent curve with the peak at the interpolation threshold. The norm of the corresponding classifier increases monotonically up to the interpolation peak and decreases beyond that. It asymptotes to the norm of the kernel machine  which can be computed using the following explicit formula  for a function written in the form of Eq.~\ref{eq:ker}) (where $K$ is the kernel matrix):
$$
\|f\|^2_{\H_K} = \balpha^{\brown{T}} K \balpha
$$

\blue{
\subsection{When do minimum norm predictors generalize?}
As we have discussed above, 
considerations of smoothness and simplicity suggest that minimum norm solutions may have favorable generalization properties.
This  turns out to be true even when the norm does not have a clear interpretation as a smoothness functional.
Indeed, consider an ostensibly simple classical regression setup, where  data satisfy a linear relation corrupted by noise $\epsilon_i$
\begin{equation}\label{eq:lin_reg}
y_i = \langle \bbeta^*,\bx_i\rangle +\epsilon_i, ~\bbeta^*\in \R^d,~ \epsilon_i \in \R,~ i=1,\ldots,n \end{equation}
In the over-parameterized setting, when $d>n$, least square regression yields a minimum norm interpolator  given by $y(\bx)=\langle\bbeta_\text{int},\bx\rangle$, where
\begin{equation}\label{eq:lin_interp}
\bbeta_\text{int} = \arg\min_{\bbeta \in \R^d,  ~ \langle \bbeta,\bx_i\rangle = y_i,~ i=1,\ldots,n} \|\bbeta\|
\end{equation}
$\bbeta_\text{int}$ can be written explicitly 
as 
$$
\bbeta_\text{int} = {\bf X}^\dagger \by 
$$
where $\bf X$ is the data matrix, $\by$ is the vector of labels and ${\bf X}^\dagger$ is the Moore-Penrose (pseudo-)inverse\footnote{\blue{If $\rmX\rmX^T$ is invertible, as is usually the case in over-parameterized settings, $\rmX^\dagger = \rmX^T(\rmX\rmX^T)^{-1}$. In contrast, if $\rmX^T\rmX$ is invertible (under the classical under-parameterized setting), $\rmX^\dagger=(\rmX^T\rmX)^{-1}\rmX^T$. Note that both  $\rmX\rmX^T$ and $\rmX^T\rmX$ matrices cannot be invertible unless $X$ is a square matrix, which occurs at the interpolation threshold.}}.
Linear regression for models of the type in Eq.~\ref{eq:lin_reg} is no doubt the oldest\footnote{\blue{Originally introduced  by Gauss and, possibly later, Legendre! See~\cite{Gauss_least_squares}.}} and best studied family of statistical methods. Yet, strikingly,  predictors such as those in  Eq.~\ref{eq:lin_interp}, have historically been mostly overlooked, at least for noisy data. 
Indeed, a classical prescription  is to {\it regularize} the predictor by, e.g., adding a ``ridge'' $\lambda I$ to obtain a non-interpolating predictor.
}
\blue{
The reluctance to overfit inhibited exploration of a range of settings where $y(\bx)=\langle \bbeta_\text{int},\bx\rangle$ provided optimal or near-optimal predictions. Very recently, these  ``harmless interpolation''~\cite{muthukumar2020harmless} or ``benign over-fitting''~\cite{bartlett2020benign} regimes have become a very active direction of research, a development inspired by efforts to understand deep learning. In particular, the work~\cite{bartlett2020benign} provided a spectral characterization of models exhibiting this behavior. In addition to the aforementioned papers, some of the first work toward understanding ``benign overfitting'' and double descent under various linear settings include~\cite{hastie2019surprises, belkin2020two, mitra2019understanding, xu2019number}. 
}\brown{Importantly, they demonstrate that when the number of parameters varies, even for linear models over-parametrized predictors are sometimes preferable to any ``classical'' under-parameterized model.}  

\blue{
Notably, even in cases when the norm clearly corresponds to measures of functional smoothness, such as the cases of RKHS or, closely related random feature maps, the analyses of interpolation for noisy data are subtle and have only recently started to appear, e.g.,~\cite{liang2020just,mei2019generalization}.  
For a far more detailed overview of the progress on interpolation in linear regression and kernel methods see the parallel Acta Numerica paper~\cite{bartlett2021deep}.
}

\blue{
\subsection{Alignment of generalization and optimization in linear and kernel models}
While over-parameterized models have manifolds of interpolating solutions, minimum norm solutions, as we have discussed,  have special properties which may be conducive to generalization. 
For over-parameterized linear and kernel models\footnote{\footnote{Kernel models are linear from the optimization point of view as they can be viewed as a fixed feature map followed by a linear method. Thus we will not make a distinction in the optimization context.  They are, however, non-linear functions of the input space.}} there is a beautiful alignment of optimization and
minimum norm interpolation: gradient descent \brown{GD} or Stochastic Gradient Descent (SGD) initialized at the origin can be guaranteed to converge to $\bbeta_\text{int}$ defined in Eq.~\ref{eq:lin_interp}.
} 
\blue{
To see why this is the case we make the following  observations:
\begin{itemize}
    \item $\bbeta_\text{int} \in \T$, where $\T=\mathop{Span}{\{x_1,\ldots,x_n\}}$ is the 
    span of the training examples (or their feature embeddings in the kernel case). To see that, verify that if $\bbeta_\text{int} \notin \T$, orthogonal projection of $\bbeta_\text{int}$ onto $\T$ is an interpolating predictor with even smaller norm, a contradiction to the definition of $\bbeta_\text{int}$.
    \item The (affine) subspace of interpolating predictors $\S$ (Eq.~\ref{eq:interp_subset}) is orthogonal to $\T$  and hence $\{\bbeta_\text{int}\} = \S \cap \T$. 
\end{itemize}
These two points together are in fact a version of the Representer theorem briefly discussed in Sec.~\ref{sec:rff}.
}

\blue{
Consider now gradient descent for linear regression  initialized at within the span of training examples $\bbeta_0 \in \T$. Typically, we simply choose $\bbeta_0 =0$ as the origin has the notable  property of  belonging to the span of any vectors. It can be easily verified that  the gradient of the loss function at any point is also in the span of the training examples and thus the whole {\it optimization path} lies within $\T$. As the gradient descent converges to a minimizer of the loss function, and $\T$ is a closed set, GD must converge to the minimum norm solution  $\bbeta_\text{int}$. Remarkably, in the over-parameterized settings convergence to $\bbeta_\text{int}$ is true for SGD, even with a fixed learning rate (see Sec.~\ref{sec:sgd}). In  contrast, under-parameterized SGD with a fixed learning rate does not converge at all.
}

\subsection{Is deep learning kernel learning? Transition to linearity in wide neural networks.}\label{sec:ntk}

But how do these ideas apply to deep neural networks? Why are complicated non-linear systems with large numbers of parameters
able to generalize to unseen data? 

It is important to recognize  that generalization in large neural networks is a robust pattern that holds across multiple dimensions of architectures, optimization methods and datasets\footnote{While details such as selection of activation functions, initialization methods, connectivity patterns or many specific parameters of training 
(annealing schedules, momentum, batch normalization, dropout, the list goes on ad infinitum), matter for state-of-the-art performance, they are almost irrelevant if the goal is to simply obtain passable generalization.
}. As such, the ability of neural networks to generalize to unseen data reflects a fundamental interaction between the mathematical structures underlying neural function spaces, algorithms and the nature of our data. 
It can be likened to the gravitational force holding the Solar System, not a momentary alignment of the planets. 

This point of view implies that understanding generalization in complex neural networks has to involve a general principle, relating them to more tractable mathematical objects. A prominent candidate  for such an object are kernel machines and their corresponding Reproducing Kernel Hilbert Spaces. As we discussed above,  Random Fourier Features-based networks, a rather specialized type of neural architectures, approximate Gaussian kernel machines.
Perhaps general neural networks can also be tied to kernel machines?  Strikingly, it turns out to be the case indeed,  at least for some classes of neural networks. 

One of the most intriguing and remarkable recent mathematical discoveries in deep learning is the constancy of the so-called Neural Tangent Kernel (NTK) for certain wide neural networks due to Jacot, Gabriel and Hongler~\cite{jacot2018neural}. As the width of certain networks increases to infinity, they undergo {\it transition to linearity} (using the term and following the discussion in~\cite{liu2020linearity}) and become linear functions of their parameters.  Specifically,  consider a model $f(\bw,\bx)$, where the vector $\bw\in \R^M$ represents trainable parameters. The {\it tangent kernel} at $\bw$, associated to $f$ is defined as follows: 
\begin{equation}\label{eq:ntk}
    K_{(\bx,\bz)}(\bw) := \langle \nabla_\bw f(\bw;\bx),\nabla_\bw f(\bw;\bz)\rangle, \quad \textrm{for fixed inputs }\bx, \bz \in \mathbb{R}^{d}.
\end{equation}

It is not difficult to verify that $K_{(\bx,\bz)}(\bw)$ is a positive semi-definite kernel function for any fixed $\bw$. To see that, consider the ``feature map'' $\phi_\bw: \R^d \to \R^M$ given by
$$
\phi_\bw (\bx) =  \nabla_\bw f(\bw;\bx)
$$
 Eq.~\ref{eq:ntk} states that the tangent kernel is simply the linear kernel in the embedding space $\R^M$, $ K_{(\bx,\bz)}(\bw) =  \langle \phi_\bw (\bx),\phi_\bw (\bz)\rangle$.   
 
 The surprising and singular finding of~\cite{jacot2018neural} 
 is that for a range of  infinitely wide neural network architectures with linear \brown{output} layer, $\phi_\bw (\blue{\bx})$ is independent of $\bw$ in a ball around a random ``initialization'' point $\bw_0$. That can be shown to be equivalent to the linearity of $f(\bw,\bx)$ in $\bw$ (and hence transition to linearity in the limit of infinite width):
 $$
 f(\bw,\bx) = \langle \bw - \bw_0,\phi_{\bw_0}(\bx)\rangle + f(\bw_0,\bx)
 $$
Note that $f(\bw,\bx)$ is not a linear predictor in $\bx$, it is a kernel machine, linear in terms of the parameter vector $\bw\in \R^M$. Importantly, $f(\bw,\bx)$ has linear training dynamics and that is the way this phenomenon is usually described in the machine learning literature (e.g.,~\cite{lee2019wide}) . However the linearity itself is a property of the model unrelated to any training procedure\footnote{This is a slight simplification as for any finite width the linearity is only approximate in a ball of a finite radius. Thus the optimization target must be contained in that ball. For the square loss it is always the case for sufficiently wide network. For cross-entropy loss it is not generally the case, see \orange{Section~\ref{sec:square_loss_class}}. }.

 To understand the nature of this transition to linearity consider the Taylor expansion of $f(\bw,\bx)$ around $\bw_0$  with the Lagrange remainder term in a ball ${\cal B}\green{\subset \R^M}$ of radius $R$ around $\bw_0$. \blue{For any $\bw \in {\cal B}$ there is $\xi \in  {\cal B}$ so that
 $$
 f(\bw,\bx) = f(\bw_0,\bx) + \langle \bw - \bw_0,\phi_{\bw_0}(\bx)\rangle + \frac{1}{2}\langle \bw-\bw_0, H(\xi) (\bw-\bw_0)\rangle
 $$}
 
We see that the deviation from the linearity is bounded by the spectral norm of the Hessian:
$$
 \sup_{\bw \in {\cal B}} f(\bw,\bx) - f(\bw_0,\bx) - \langle \bw - \bw_0,\phi_{\bw_0}(\bx)\rangle \le \frac{\green{R^2}}{2} \sup_{\xi \in {\cal B}} \|H(\xi)\|
$$
 
A general (feed-forward) neural network with $L$ hidden layers and a linear output layer is a function defined recursively as:
\begin{align}\label{eq:generalnn}
 &\balpha^{(0)} = \bx, \nonumber\\
 &\balpha^{(l)}= \phi_{l}(\bW^{(l)}*\balpha^{(l-1)}), \balpha \in \R^{d_l}, \bW^{(l)} \in \R^{d_{l}\times d_{l-1}},  \ l = 1,2,\ldots, L,\nonumber\\
 &f(\bw,\bx) = \frac{1}{\sqrt{m}}\bv^T \balpha^{(L)}, \bv \in \R^{d_L}
\end{align}
The parameter vector $\bw$ is obtained by concatenation of all weight vectors 
$\bw = (\bw^{(1)},\ldots,\bw^{(L)}, \bv)$ and the {\it activation functions}
 $\phi_l$  are usually applied coordinate-wise.  It turns out these, seemingly complex, non-linear systems  exhibit transition to linearity under quite general conditions (see~\cite{liu2020linearity}), given appropriate random initialization $\bw_0$. Specifically, it can be shown that for a ball $\cal B$ of fixed radius  around the initialization \brown{$\bw_0$} \blue{the spectral norm of the Hessian satisfies}
 \begin{equation}\label{eq:sqrtm}
 \sup_{\xi \in \cal B} \|H(\xi)\| \le O^*\left(\frac{\orange{1}}{\sqrt m}\right), ~\text{where } m =\min_{l=1, \ldots, L}(d_l) 
\end{equation}
It is important to emphasize that linearity is a true emerging property of large systems and does not come from the scaling of the function value with the increasing  width $m$. Indeed, for any $m$ the value of the function at initialization and its gradient are all of order $1$, $f(\bw,x) = \Omega(1)$, $\nabla f(\bw,x) = \Omega(1)$. 
\blue{
\paragraph{Two-layer network: an illustration.}
To provide some intuition for this structural phenomenon consider a particularly simple case of a  two-layer neural network with fixed second layer. 
Let the model $f(\bw, x), x\in\R$ be of the form
\begin{equation}\label{eq:2layer}
    f(\bw,x) = \frac{1}{\sqrt{m}}\sum_{i=1}^m v_i \alpha(w_i x),
\end{equation}
For simplicity, assume that  $v_i\in \{-1,1\}$ are {\it fixed} and $w_i$ are trainable parameters. 
It is easy to see that in this case the Hessian $H(\bw)$ is a diagonal matrix with 
\begin{equation}\label{eq:hessian_simple_model}
    (H)_{ii} = \frac{1}{\sqrt{m}} v_i\frac{d^2\alpha(w_i\, x)}{d^2 w_i} =  \pm\frac{1}{\sqrt{m}} x^2\alpha''(w_i x) .
\end{equation}
We see that 
$$
\|H(\bw)\| = \frac{x^2}{\sqrt{m}} \max_i |\alpha''(w_ix)| = 
\frac{x^2}{\sqrt{m}} \|\underbrace{(\alpha''(w_1x),\ldots,\alpha''(w_mx))}_{\ba}\|_\infty
$$
In contrast, the tangent kernel
$$
\|\nabla_\bw f\| = \sqrt{\frac{1}{m} \sum_i x^2 (\alpha'(w_ix))^2} = \frac{x}{\sqrt{m}} \|\underbrace{(\alpha'(w_1x),\ldots,\alpha'(w_mx))}_{\bb}\|
$$
Assuming that $\bw$ is such, that
$\alpha'(w_ix)$ and $\alpha''(w_jx)$ are of all of the same order, from the relationship between 2-norm and $\infty$-norm in $\R^m$ we expect  
$$ \|\bb\| \sim \sqrt{m}\,\|\ba\|_\infty.$$
Hence, 
$$
\|H(\bw)\| \sim \frac{1}{\sqrt{m}} \,\|\nabla_\bw f\|
$$
Thus, we see that the structure of the Hessian matrix forces its spectral norm to be a factor of $\sqrt{m}$\,  smaller compared to the gradient. 
If (following a common practice) $w_i$ are sampled iid from the standard normal distribution 
\begin{equation}\label{eq:scaling}
\|\nabla_\bw f\| = \sqrt{K_{(\bw,\bw)}(x)} = \Omega(1),~~\|H(\bw)\| = O\left(\frac{1}{\sqrt{m}}\right)
\end{equation}
If, furthermore, the second layer weights $v_i$ are sampled with expected value zero, $f(\bw,x)=O(1)$.
Note that to ensure the transition to linearity we need for the scaling in Eq.~\ref{eq:scaling} to hold in ball of radius $O(1)$ around $\bw$ (rather than just at the point $\bw$), which, in this case,  can be easily verified. 
}

\blue{
The example above illustrates how the transition to linearity is the result of the structural properties of the network (in this case the Hessian is a diagonal matrix) and the difference between the $2$-norm ind $\infty$-norm in a high-dimensional space. For general deep networks the Hessian is no longer diagonal, and the argument is more involved,  yet there is a similar structural difference between the gradient and the Hessian related to different scaling of the $2$ and $\infty$ norms with dimension. 
}

Furthermore, transition to linearity is not simply a property of large systems. Indeed, adding a non-linearity at the output layer, i.e., defining 
\blue{
$$
g(\bw,x) = \phi(f(\bw,x))
$$
where $f(\bw,x)$ is defined by Eq.~\ref{eq:2layer} and $\phi$ is any smooth  function with non-zero second derivative breaks the transition to linearity  independently of the width $m$ and the function  $\phi$. To see that, observe that the Hessian of $g$, $H_g$ can be written, in terms of the gradient and Hessian of $f$, ($\nabla_\bw f$ and $H(\bw)$, respectively) as   
\begin{equation}\label{eq:notransition}
H_g(\bw) =   \phi'(f)\underbrace{ H(\bw)}_{O(1/\sqrt{m})} +\phi''(f)\, \underbrace{\nabla_\bw f \times (\nabla_\bw f)^T}_{\Omega(1)}
\end{equation}
We see that the second term in Eq.~\ref{eq:notransition} is of the order $\|\nabla_\bw f\|^2=\Omega(1)$ and does not scale with $m$. Thus the transition to linearity does not occur and the tangent kernel does not become constant in a ball of a fixed radius even as the width of the network tends to infinity.  
}
\blue{Interestingly,} introducing even a single narrow ``bottleneck'' layer has the same effect even if the activation functions in that layer are linear (as long as some activation functions in at least one of the deeper layers are non-linear). 

As we will discuss later in Section~\ref{sec:conv}, the \green{transition to linearity} is not needed for optimization, which makes this phenomenon even more intriguing. Indeed, it is possible to imagine a world where the transition to linearity phenomenon does not exist, yet neural networks  can still be optimized using the usual gradient-based methods. 

It is thus even more fascinating that a large class of very
complex functions turn out to be linear in parameters and the corresponding complex learning algorithms are simply training kernel machines. 
In my view this adds significantly to the evidence that {\it understanding kernel learning is a key to deep learning} as we argued in~\cite{belkin2018understand}. 
Some important caveats are in order. 
While it is arguable that deep learning may be equivalent to kernel learning in some interesting and practical regimes, the jury is still out on the question of whether this point of view can provide a conclusive understanding of generalization in neural networks. Indeed a considerable amount of recent theoretical work has been aimed at trying to understand regimes (sometimes called the ``rich regimes'', e.g., ~\cite{woodworth2020kernel,ghorbani2020neural}) where the transition to linearity does not happen and the system is non-linear throughout the training process. Other work (going back to~\cite{warmuth2005leaving}) argues that there are theoretical barriers separating function classes learnable by neural networks and kernel machines~\cite{allen2020backward,pravesh2020expressive}. 
Whether these analyses are relevant for explaining empirically observed behaviours of deep networks still requires further exploration. 

Please also see some discussion of these issues in Section~\ref{sec:through}.

\section{\brown{The wonders of optimization}} \label{sec:conv}


The  success of deep learning has heavily relied on the remarkable effectiveness of gradient-based \brown{optimization} methods, such as stochastic gradient descent (SGD), applied to large non-linear neural networks. \brown{Classically, finding global minima in non-convex problems, such as these, has been considered intractable and yet, in practice, neural networks can be reliably trained.
}

\brown{Over-parameterization
and interpolation provide a  distinct perspective on optimization. Under-parameterized problems are typically locally convex around their local minima. In contrast, over-parameterized  non-linear optimization landscapes are \green{generically} non-convex, even locally. Instead, as we will argue, throughout most \green{(but not all)} of the parameter space they satisfy the Polyak\,-\,{\L}ojasiewicz condition, which guarantees both existence of global minima within any sufficiently large ball and convergence of gradient methods, including GD and SGD.}

\brown{Finally, as we discuss in Sec.~\ref{sec:sgd}, interpolation sheds light on a separate empirically observed phenomenon,
the striking effectiveness of mini-batch SGD (ubiquitous in applications) in comparison to the standard gradient descent.}

\subsection{\brown{From convexity to the PL* condition}}
Mathematically, interpolation corresponds to identifying $\bw$ so that $$f(\bw, \bx_i)=y_i, i=1,\ldots,n, \bx_i \in \R^d, \bw\in \R^M.$$ This is a system of $n$ equations with $M$ variables. Aggregating these equations into a single map,
\begin{equation}\label{eq:main_opt}
F(\bw)=(f(\bw,\bx_1),\ldots, f(\bw,\bx_n)\blue{)},
\end{equation}
and setting $\by=(y_1,\ldots,y_n)$,
we can write that $\bw$ is a solution for a single equation 
\begin{equation}\label{eq:opt}
F(\bw) = \by, ~~F: \mathbb{R}^M \to \R^n.
\end{equation}

When can such a system be solved? The question posed in such generality initially appears to be absurd.  A special case, that of solving systems of polynomial equations, is at the core of algebraic geometry, a deep and intricate mathematical field. 
And yet, we can often easily train non-linear neural networks to fit arbitrary data~\cite{zhang2016understanding}. Furthermore, practical neural networks are typically trained using simple first order gradient-based methods,  such as stochastic gradient descent (SGD).

\begin{figure}
    \centering
    \begin{subfigure}[t]{0.45\textwidth}
        \centering
        \includegraphics[width=\linewidth]{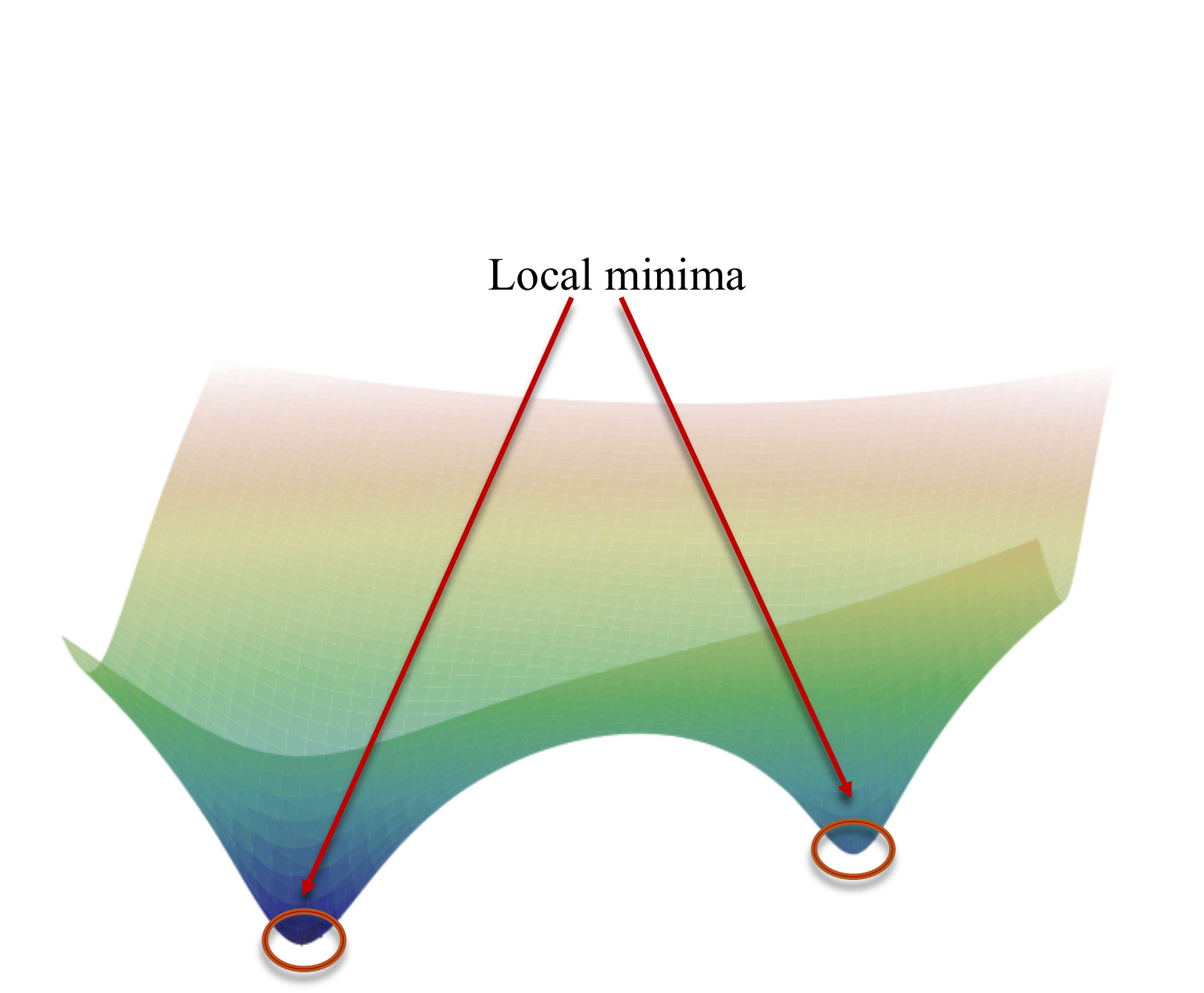} 
        \caption{Under-parameterized models} \label{fig:loc}
    \end{subfigure}
    \hfill
    \begin{subfigure}[t]{0.45\textwidth}
        \centering
        \includegraphics[width=\linewidth]{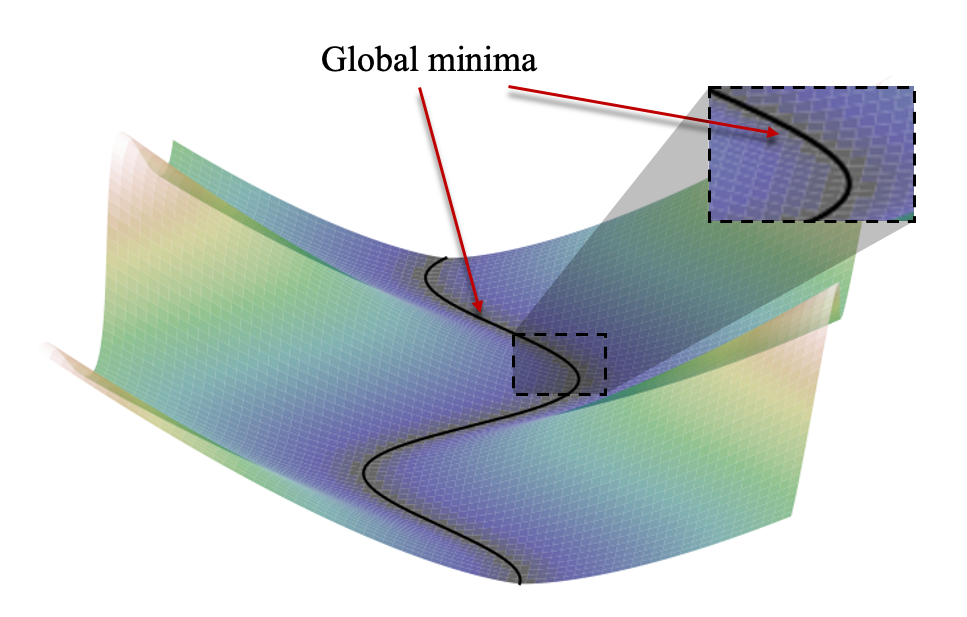} 
        \caption{Over-parameterized models} \label{fig:glob}
    \end{subfigure}
     \caption{Panel (a): Loss landscape is locally convex at local minima. Panel (b): Loss landscape is incompatible with local convexity when the set of global minima is not linear (insert). Figure credit:~\cite{liu2020toward}. }\label{fig:loc&glob}
  \end{figure}

The idea of over-parameterization has recently emerged as an explanation for this phenomenon based on the intuition that a system with more variables than equations can generically be solved.
We first observe that solving Eq.~\ref{eq:opt} (assuming a solution exists) is equivalent to minimizing the loss function 
$$
\LL(\bw) = \|F(\bw)-\by\|^2.
$$
\blue{This is a non-linear least squares problem, which is well-studied under classical under-parameterized settings (see~\cite{nocedal2006numerical}, Chapter 10).
What property of the over-parameterized optimization landscape allows for effective optimization by gradient descent (GD) or its variants? }
It is instructive to consider a simple example in Fig.~\ref{fig:loc&glob} (from~\cite{liu2020toward}). 
The left panel corresponds to the classical regime with many 
 isolated local minima.  We see that for such a landscape there is little hope that a local method, such as GD can reach a global optimum. 
Instead we expect it to converge to a local minimum close to the initialization point. Note that in a neighborhood of a local \blue{minimizer} the function is convex and classical convergence analyses apply. 

A key insight is that landscapes of over-parameterized systems look very differently, like the right panel in Fig~\ref{fig:glob}. We see that there every local minimum is global and the manifold of minimizers $\S$ has positive dimension. It is important to observe that such a landscape is incompatible with convexity even locally. Indeed, consider an arbitrary point  $s \in \S$
inside the insert in Fig~\ref{fig:glob}.
If $\LL(\bw)$ is convex in any ball  ${\cal B} \subset \S$ around $s$, the set of minimizers within that neighborhood,  ${\cal B} \cap \S$ must be a a convex set in $\R^M$.  Hence $\S$ must be a locally linear manifold near $s$ for $\LL$ to be locally convex. It is, of course, not the case for general systems and  cannot be  expected, even at a single point. 

Thus, one of the key lessons of deep learning in optimization:\\ 
{\it Convexity, even locally, cannot be the basis of analysis for over-parameterized systems. }

But what mathematical property encapsulates ability to optimize by gradient descent for landscapes, such as in Fig.~\ref{fig:loc&glob}. 
It turns out that a simple condition proposed in 1963 by Polyak~\cite{polyak1963gradient} is  sufficient for efficient minimization by gradient descent. This PL-condition 
(for Polyak and also {\L}ojasiewicz, who independently analyzed a more general version of the condition  in a different context~\cite{lojasiewicz1963topological}) is a simple first order inequality applicable to a broad range of optimization problems~\cite{karimi2016linear}.  

We say that $\LL(\bw)$ is $\mu$-PL, if the following holds:
\begin{equation}\label{eq:PL}
    \frac{1}{2}\|\nabla \LL(\bw)\|^2 \ge \mu (\LL(\bw)-\LL(\bw^*)), 
\end{equation}
Here $\bw^*$ is a global minimizer and $\mu >0$ is a fixed real number. 
The original Polyak's work~\cite{polyak1963gradient} showed that PL condition within a sufficiently large ball (with radius $O(1/\mu)$) implied convergence of gradient descent. 

It is important to notice that, unlike convexity, PL-condition is compatible with curved manifolds of minimizers. 
However, in this formulation, the condition is non-local.
While convexity can be verified point-wise by making sure that the Hessian of \brown{$\LL$} is positive \blue{semi-}definite, the PL condition requires ''oracle'' knowledge of $\LL(\bw^*)$. This lack of point-wise verifiability is perhaps the reason PL-condition has not been used more widely in the optimization literature.

However simply removing the $L(\bw^*)$ from Eq.~\ref{eq:PL}  addresses this issue in over-parameterized settings! Consider the following modification called  PL* in~\cite{liu2020toward} and local PL in~\cite{oymak2019overparameterized}.
\begin{equation*}
    \frac{1}{2}\|\nabla \LL(\bw)\|^2 \ge \mu \LL(\bw), 
\end{equation*}
It turns out that PL* condition in a ball of sufficiently large radius implies both existence of an interpolating solution within that ball and exponential convergence of gradient descent and, indeed, stochastic gradient descent. 

It is interesting to note that PL* is not a  useful concept in under-parameterized settings -- generically, there is no solution to $F(\bw)=\by$ and thus the condition cannot be satisfied \blue{along the whole optimization path}. On the other hand, the condition is remarkably flexible -- it naturally extends to Riemannian manifolds (we only need the gradient to be defined) and is invariant under non-degenerate coordinate transformations.   
 \begin{figure}[t]
        \centering
        \includegraphics[width=1\linewidth]{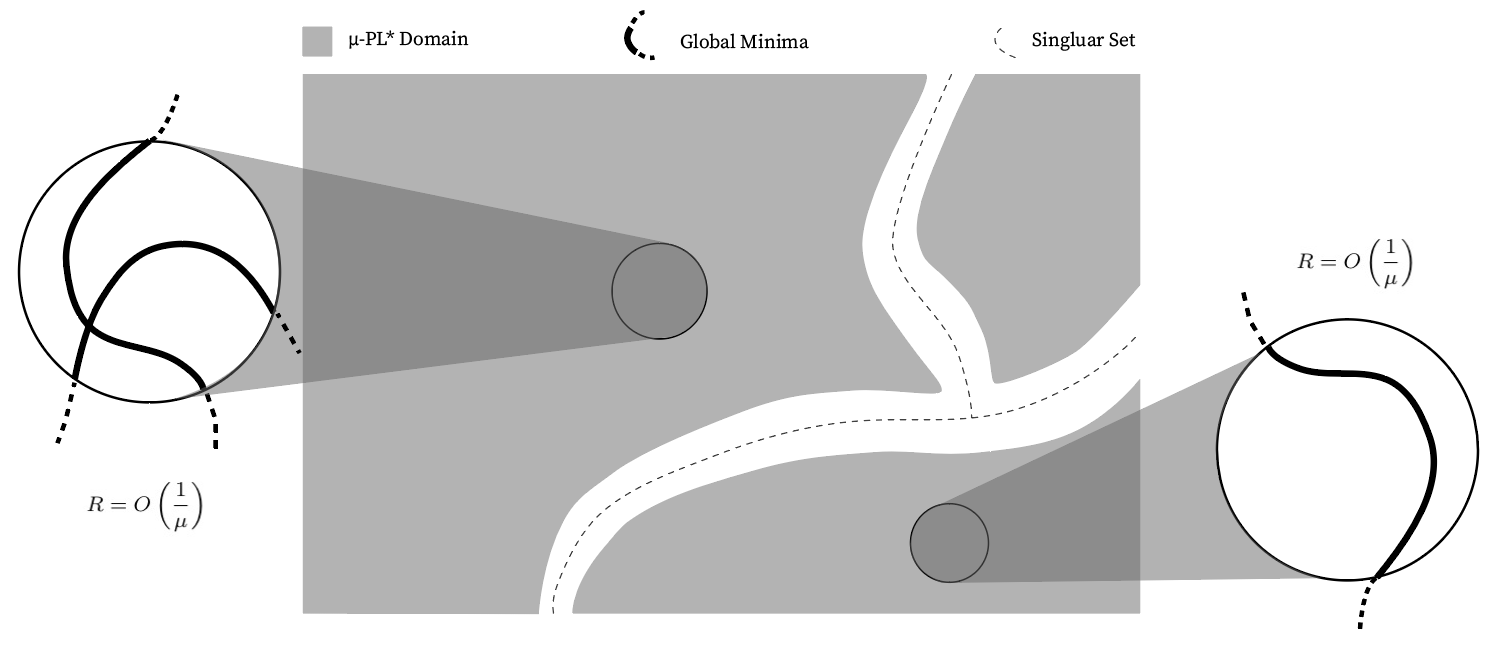} 
        \caption{\blue{
        The loss function $\LL(\bw)$ is $\mu$-PL* inside the shaded domain. Singular set correspond to parameters $\bw$ with degenerate tangent kernel $K(\bw)$. 
        Every ball of radius $O(1/\mu)$ within the shaded set intersects with the set of global minima of $\LL(\bw)$, i.e.,  solutions to $F(\bw) = \by$. Figure credit:~\cite{liu2020toward}. }} \label{fig:pldomain}
  \end{figure}

\subsection{\brown{Condition numbers of nonlinear systems}}

Why do over-parameterized systems satisfy the PL* condition? The reason is closely related to the Tangent Kernel discussed in Section~\ref{sec:ntk}. Consider the tangent kernel of the map $F(\bw)$ defined as $n\times n$ matrix valued function 
$$
K(\bw) = DF^T(\bw) \times DF(\bw), DF(\bw) \in \R^{M\times n} 
$$
where $DF$ is the differential of the map $F$. It can be shown for the square loss $\LL(\bw)$  satisfies the  PL*- condition with $\mu=\lmin(K)$. 
Note that the rank of $K$ is less or equal to $M$.
Hence, if the system is under-parameterized, i.e., $M<n$,
$\lmin(K)(\bw) \equiv 0$ and the corresponding PL* condition is always trivial. 

In contrast, when $M\ge n$, we expect $\lmin(K)(\bw) >0$ for generic $\bw$. More precisely, by parameter counting, we expect that the set of of $\bw$ with singular Tangent Kernel $\{\bw \in \R^M: \lmin(K)(\bw)= 0\}$ is of co-dimension $M-n+1$, which is exactly the amount of over-parameterization. 
Thus, we expect that large subsets of the space $\R^M$ have eigenvalues separated from zero, $\lmin(K)(\bw) \ge \mu$. This is depicted graphically in Fig.~\ref{fig:pldomain} (from~\cite{liu2020toward}). The shaded areas correspond to the sets where the loss function is \green{$\mu$-PL*}.  
In order to make sure that solution to the Eq.~\ref{eq:main_opt} exists and can be achieved by Gradient Descent, we need to make sure that $\lmin(K)(\bw)>\mu$ in a  ball of radius $O\left(\frac{1}{\mu}\right)$. Every such ball in the shaded area contains solutions of Eq.~\ref{eq:main_opt} (global minima of the loss function). 

But how can an analytic condition, like a lower bound on the smallest eigenvalue of the tangent kernel, be verified for models such as neural networks? 

\subsection{\brown{Controlling PL* condition of neural networks}}
\brown{
As discussed above and graphically illustrated in Fig.~\ref{fig:pldomain}, we expect over-parameterized systems to satisfy the PL* condition over most of the parameter space. Yet, explicitly controlling $\mu=\lmin(K)$ in a ball of a certain radius can be subtle. We can identify two techniques which help establish such control for  neural networks and other systems. The first one, the Hessian control, uses the fact that near-linear systems are well-conditioned in a ball, provided they are well-conditioned at the origin. 
The second, transformation control, is based on the observation that well-conditioned systems stay such under composition with  ``benign'' transformations.
}
\green{Combining these techniques can be used to prove convergence of randomly initialized wide neural networks.}

\subsubsection{\brown{Hessian control}}

Transition to linearity, discussed in Section~\ref{sec:ntk}, provides a powerful (if somewhat crude) tool for controlling  $\lmin(K)$ for wide networks.  
The key observation is that \green{$K(\bw)$} \green{is closely related} to the first derivative of $F$ \green{at $\bw$}. Thus the change of $K(\bw)$ 
from the initialization $K(\bw_0)$ can be bounded in terms of the norm of the Hessian $H$, the second derivative of \green{$F$} using, essentially, the mean value theorem.   We can bound the operator norm to get the following inequality \brown{(see~\cite{liu2020linearity})}:
\begin{equation}\label{eq:hessian_control1}
\forall \bw\in \B_R ~~~\|K(\bw) - K(\bw_0)\|\le 
O\left( R \max_{\xi \in \B_R}  \|H(\xi)\| \right)
\end{equation}
where $\B_R$ is a ball of radius $R$ around $\bw_0$.  

Using standard eigenvalue perturbation bounds we have
\begin{equation}\label{eq:hessian_control}
\forall \bw\in \B_R ~~~|\lmin(K)(\bw) - \lmin(K)(\bw_0)|\le 
O\left( R \max_{\xi \in \B_R}  \|H(\xi)\| \right)
\end{equation}
Recall (Eq.~\ref{eq:sqrtm}) that for networks of width $m$ with linear last layer $\|H\|=O(1/\sqrt{m})$. On the other hand, it can be shown (e.g.,~\cite{du2018gradientshallow} and~\cite{du2018gradientdeep} for shallow and deep networks respectively) that $\lmin(K)(\bw_0) = O(1)$ and is essentially independent of the width. 
\green{
\orange{Hence Eq.~\ref{eq:hessian_control} guarantees that given any fixed radius $R$, for a sufficiently wide network $\lmin(K)(\bw)$ is separated from zero in the ball $\B_R$.  Thus the loss function satisfies the PL* condition in $\B_R$.} As we discussed above, this guarantees the existence of global minima of the loss function and convergence of gradient descent for wide neural networks with linear output layer. 
}

\subsubsection{\brown{Transformation control}}
\brown{Another way to control the condition number of a system is by representing it as a composition of
two or more well-conditioned maps. 
}

Informally, due to the chain rule, if $F$  is well conditioned, so is  $\phi \circ F \circ \psi (\bw)$, where 
$$
\phi:\R^n \to \R^n, ~~~ \psi: \R^m\to \R^m
$$
are maps with non-degenerate Jacobian matrices.

In particular, combining Hessian control with transformation control, can be used to prove convergence for wide neural networks with non-linear last layer~\cite{liu2020linearity}.

\subsection{Efficient optimization by SGD}\label{sec:sgd}
We have seen that over-parameterization helps explain why Gradient Descent  can reach global minima even for highly non-convex optimization landscapes. Yet, in practice, GD is rarely used. Instead, mini-batch stochastic methods, such as SGD or Adam~\cite{kingma2014adam} are employed almost exclusively. 
In its simplest form, mini-batch SGD uses the following update rule:
\begin{align}
\bw_{t+1}&=\bw_t -\eta \nabla\left(\frac{1}{m}\sum_{j=1}^m\ell(f(\bw_t,\bx_{i_j}), y_{i_j})\right) \label{main-update-step}
\end{align}
Here $\{(\bx_{i_1},y_{i_1}),\ldots, (\bx_{i_m},y_{i_m})\}$ is a {\it mini-batch}, a subset of the training data of size $m$, chosen at random or sequentially and $\eta>0$ is the {\it learning rate}. 

At a first glance, from a classical point of view, it appears that GD should be preferable to SGD. In a standard convex setting GD converges at an exponential (referred as {\it linear} in the optimization literature) rate, where the loss function decreases exponentially with the number of iterations. 
In contrast, while SGD requires a factor of $\frac{n}{m}$ less computation than GD per iteration, it converges at a far slower {\it sublinear} rate (see~\cite{bubeck2015convex} for a review), with the loss function decreasing
proportionally to the inverse of the number of iterations.
Variance reduction techniques~\cite{roux2012stochastic, johnson2013accelerating, defazio2014saga} can close the gap theoretically but are rarely used in practice. 

As it turns out,  interpolation can explain the  surprising effectiveness of plain  SGD compared to GD and other non-stochastic methods\footnote{Note that the analysis is for the convex interpolated setting. While bounds for convergence under the \green{PL*} condition are available~\cite{bassily2018exponential}, they do not appear to be tight in terms of the step size and hence do not show an unambiguous advantage over GD. However, empirical evidence suggests that analogous results indeed hold \green{in practice for neural networks}.}

The key observation is that in the interpolated regime SGD with fixed step size converges exponentially fast for convex loss functions. The results showing exponential convergence of SGD when the optimal solution minimizes  the loss function at each point go back to the Kaczmarz method~\cite{kaczmarz1937angenaherte} for quadratic functions, more recently analyzed in~\cite{strohmer2009randomized}. For the general convex case, it was first shown in ~\cite{moulines2011non}. The rate was later improved in~\cite{needell2014stochastic}.

\begin{figure}
  \centering
  \includegraphics[width=0.7\textwidth]{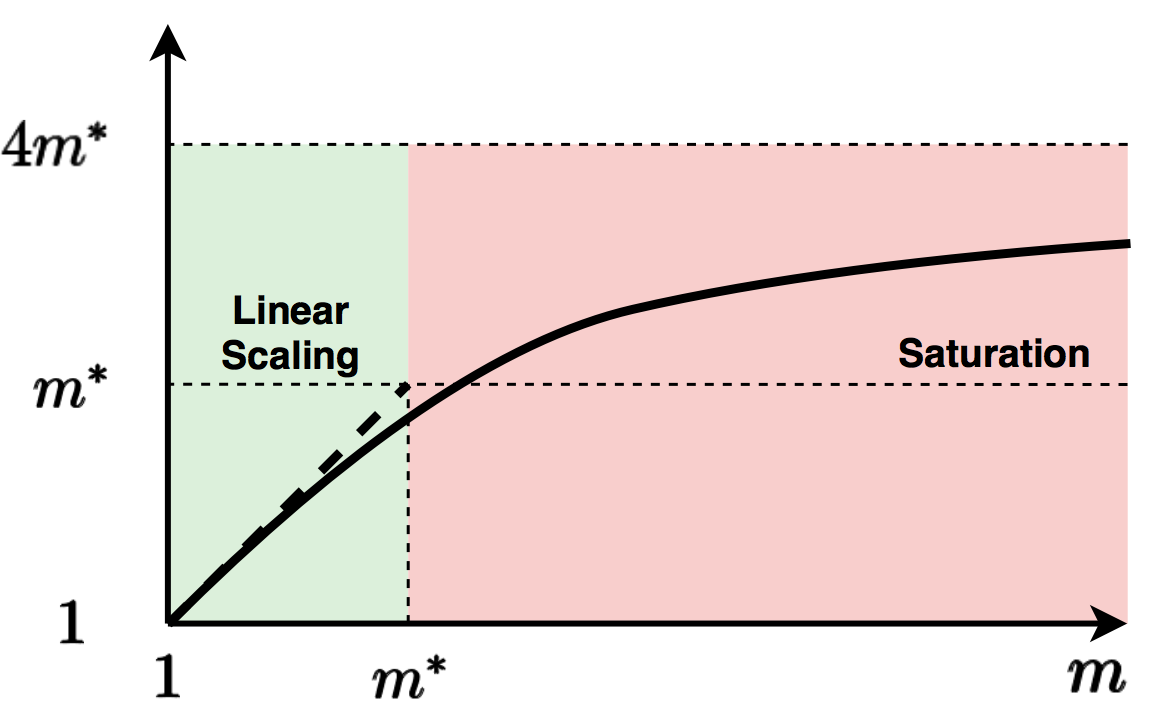}
  \caption{Number of iterations with batch size $1$ (the $y$ axis) equivalent to one iteration with batch size $m$. Critical batch size $m^*$ separates linear scaling and  regimes. Figure credit:~\cite{ma2018power}.}
  \label{fig:optimal-region-0}
\end{figure}

Intuitively, exponential convergence of SGD under interpolation is due to what may be termed ``automatic variance reduction''(\cite{liu2018accelerating}). As we approach interpolation, the loss  at every data point nears zero, and the variance due to mini-batch selection decreases accordingly. In contrast, under classical under-parameterized settings, it is impossible to satisfy all of the constraints at once, and 
the mini-batch variance converges to a non-zero constant. Thus SGD will not converge without additional algorithmic ingredients,  such as averaging or reducing the learning rate.
However, exponential convergence on its own is not enough to explain the apparent empirical superiority of SGD. An analysis in~\cite{ma2018power}, identifies  interpolation as the key to efficiency of SGD in modern ML, and provides a sharp computational characterization of the advantage in the convex case.  As the mini-batch size $m$ grows, there are two distinct regimes, separated by the {\it critical value $m^*$}: 
\begin{itemize}
    \item Linear scaling: One  SGD iteration with mini-batch of size $m\le m^*$ is  equivalent to $m$ iterations  of mini-batch of size one up to a multiplicative constant close to $1$.
    \item (saturation) One SGD iterations with a mini-batch of size $m > m^*$ is   as effective (up to a small multiplicative  constant) as one iteration of SGD with mini-batch $m^*$ or as one iteration of full gradient descent. 
\end{itemize}
For the quadratic model, $m^*= \frac{\max_{i=1}^n\{{\|\bx_i\|}^2\}}{\lambda_{max} (H)}  \le \frac{\tr(H)}{\lambda_{max} (H)} $
, where $H$ is the Hessian of the loss function and $\lambda_{max}$ is its largest eigenvalue.  This dependence is graphically represented in Fig.~\ref{fig:optimal-region-0} from~\cite{ma2018power}.

Thus, we see that the computational  savings of SGD with mini-batch size smaller than the critical size $m^*$ over GD  are of the order $\frac{n}{m^*}\approx n \frac{\lambda_{max}(H)}{\tr(H)}$. In practice, at least for kernel methods $m^*$ appears to be a small number,  less than $100$~\cite{ma2018power}. It is important to note that $m^*$ is essentially independent on $n$ -- we expect it to converge to a constant as $n\to\infty$. Thus, small (below the critical batch size)  mini-batch SGD, has $O(n)$ computational advantage over GD. 

To give a simple realistic  example, if $n=10^6$ and $m^*=10$, SGD has a factor of $~10^5$ advantage over GD, a truly remarkable improvement!

\section{\brown{Odds and ends}}\label{sec:square_loss}

\subsection{\brown{Square loss for training in classification?}} \label{sec:square_loss_class}
The attentive reader will note that most of our optimization discussions (as well as in much of  the literature)  involved the square loss. 
While training using the square loss is standard for regression tasks, it is rarely employed for classification, where  the cross-entropy loss function is the standard choice for training. For two class problems with labels $y_i \in \{1,-1\}$ the cross-entropy (logistic) loss function is defined as 
\begin{equation}
  l_{ce}(f(\bx_i),y_i) = \log\left(1+e^{-y_if(\bx_i)}\right)
\end{equation}

A striking aspect of cross-entropy is that in order to achieve zero loss we need to have $y_if(\bx_i) = \infty$. 
Thus, interpolation only occurs at infinity and any optimization procedure would eventually escape from a ball of any fixed radius. This presents difficulties for optimization analysis as it is typically harder to apply at infinity. Furthermore, since the norm of the solution vector is infinite, 
there can be no transition to linearity on any domain that includes the whole optimization path, no matter how wide  our network is and how tightly we control the Hessian norm (see Section~\ref{sec:ntk}). \brown{Finally, analyses of cross-entropy  in the linear case~\cite{ji2019implicit} suggest that  convergence is much slower than for the square loss and thus we are unlikely to approach interpolation in practice.}

Thus the use of the cross-entropy loss leads us away from interpolating solutions and toward more complex mathematical analyses. 
Does the prism of interpolation fail us at this junction?  

The accepted justification of the cross-entropy loss for classification is that it is a better ``surrogate'' for the 0-1 classification loss than the square loss (e.g.,~\citep{Goodfellow-et-al-2016}, Section 8.1.2). There is little theoretical analysis supporting this point of view. 
To the contrary,  very recent theoretical works~\citep{mai2019high, muthukumar2020classification,thrampoulidis2020theoretical} prove that in certain over-parameterized regimes, training using the square loss for classification is at least as good or better than using other loss functions. 
Furthermore, extensive empirical evaluations conducted   in~\cite{hui2021evaluation} show that modern neural architectures trained with the square loss slightly  outperform same architectures trained with the cross-entropy loss on the majority of tasks across several application domains including Natural Language Processing, Speech Recognition and Computer Vision.

A curious historical parallel is that  current reliance on cross-entropy loss in classification reminiscent of the predominance of the hinge loss in the era of the Support Vector Machines (SVM). At the time, the prevailing intuition had been that the hinge loss was preferable to the  square loss for training classifiers. Yet, the empirical evidence  had been decidedly mixed. In his remarkable 2002 thesis~\citep{rifkin2002everything}, Ryan Rifkin conducted an extensive empirical evaluation and concluded that {\it ``the performance of the RLSC} [square loss] {\it is essentially equivalent to that of the SVM} [hinge loss]  {\it across a wide range of problems, and the choice between the two should be based on computational tractability considerations''}. 

We see that interpolation as a guiding principle  points us in a right direction yet again. Furthermore, by suggesting the square loss for classification, it reveals shortcomings of theoretical intuitions and the pitfalls of excessive belief in empirical best practices.

\subsection{\brown{Interpolation and adversarial examples}}
\label{sec:adversarial}

A remarkable feature of modern neural networks is existence  of {\it adversarial examples.}
It was observed in~\cite{szegedy2013adversarial} that by adding a small, visually imperceptible, perturbation   of the pixels, an image correctly classified as ``dog'' can be moved to class ``ostrich'' or to some other obviously visually incorrect class. Far from being an isolated curiosity, this turned out to be a robust and ubiquitous property among  different neural architectures. Indeed, modifying a single, carefully selected, pixel is frequently enough to coax a neural net into misclassifying an image~\cite{su2017one}.

\begin{figure}
  \begin{center}
    \includegraphics[width=.7\textwidth]{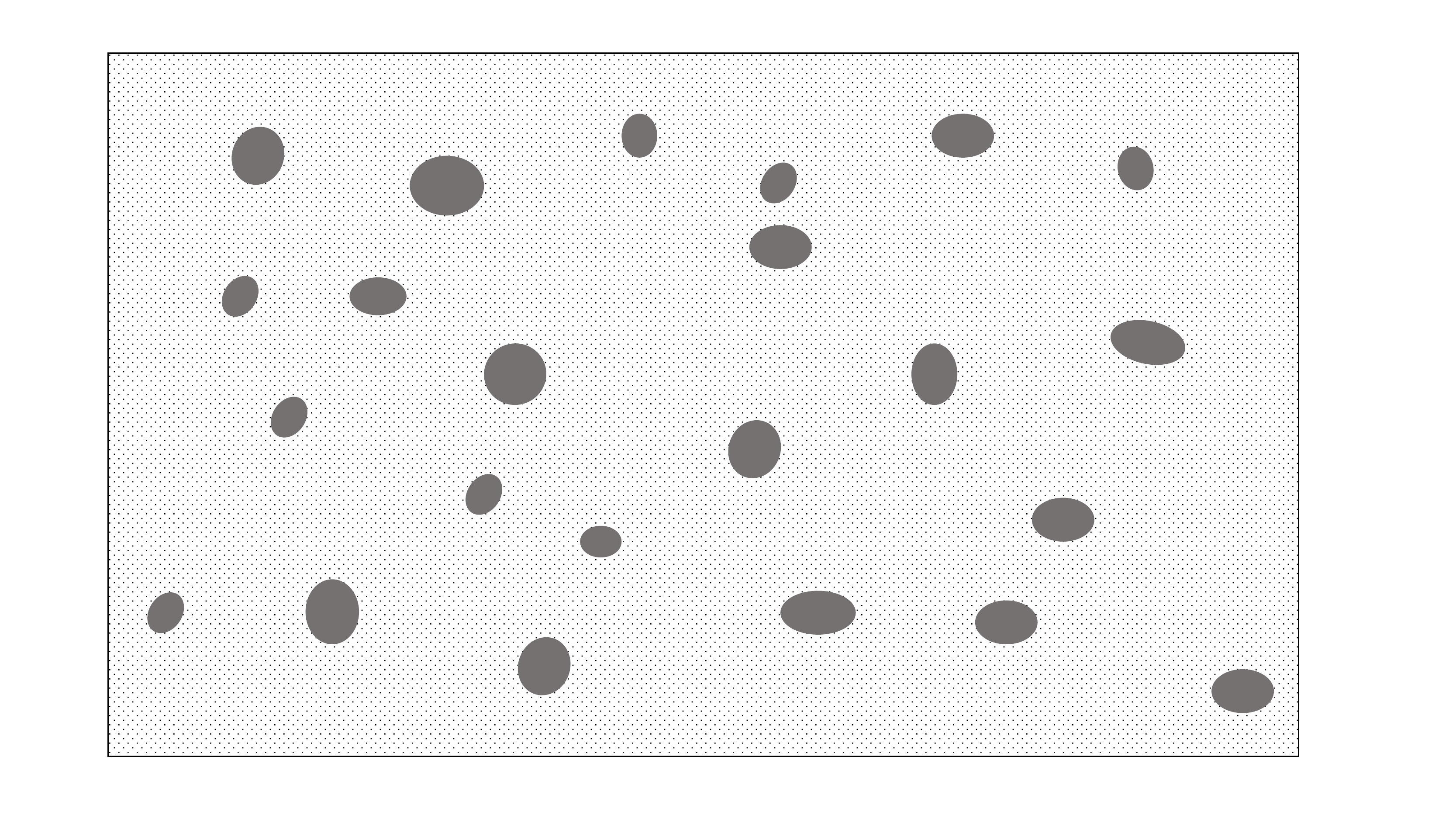}
  \end{center}
  \caption{
  Raisin bread: The ``raisins'' are basins where the interpolating predictor $\fint$ disagrees with the optimal predictor $f^*$,  surrounding ``noisy'' data points. The union of basins is an everywhere dense set of zero measure (as $n\to \infty$). }
\end{figure}
The full implications and mechanisms for the emergence of adversarial examples are not yet fully understood and are an active area of research. \blue{Among other things, the existence and pervasiveness of adversarial examples points to the limitations of the standard iid models as these data are not sampled from the same distribution as the training set.} 
Yet, it can be proved mathematically that adversarial examples are unavoidable for interpolating classifiers in the presence of label noise~\cite{belkin2018overfitting} (Theorem~5.1).
Specifically, suppose $\fint$ is an interpolating classifier and let $\bx$ be an arbitrary point. Assume that $\fint(\bx)=y$ is a correct prediction. 
Given a sufficiently large dataset, there will be at least one ''noisy'' point $\bx_i, y_i,$, such as $\fopt(\bx_i)\ne y_i$, in a small neighborhood of $\bx$ and thus a small perturbation of $\bx$ can be used to flip the label.     

If, furthermore, $\fint$ is a consistent classifier, such as  predictors discussed in Section~\ref{sec:singular_ker}, it will approach the optimal predictor $\fopt$ as the data size grows. 

Specifically, consider the set where predictions of $\fint$ differ from the optimal classification
$$
\S_n=\{\bx:\fopt(\bx)\ne\fint(\bx)\}
$$
From consistency, we have 
$$
\lim_{n\to \infty} \mu(\S_n) =0
$$
where $\mu$ is marginal probability measure of the data distribution. 
On the other hand, as $n\to \infty$,  $S_n$ becomes a dense subset of the data domain. 
This can be thought of as a raisin bread\footnote{Any similarity to the ``plum pudding'' model of the atom due to J.J.Thompson is purely coincidental.}. The are  the incorrect classification basins around each  misclassified example, i.e., the areas where the output of $\fint$ differs from $\fopt$. While the seeds permeate the bread, they occupy negligible volume inside.

This picture is indeed consistent with the extensive empirical evidence for neural networks. A random perturbation avoids adversarial ``raisins''~\cite{fawzi2016robustness}, yet they are easy to find by targeted optimization methods such as PCG~\cite{madry2017towards}. I should point out that there are also other explanations for adversarial examples~\cite{ilyas2019adversarial}. It seems plausible that several mathematical effects combine to produce adversarial examples. 

\section{\brown{Summary and  thoughts}}
\brown{
We proceed to summarize the key points of this article and conclude with a discussion of machine learning and some key questions still unresolved. 
}

\subsection{\blue{The two regimes of machine learning}}

\green{
 The sharp contrast between the ``classical'' and ``modern'' regimes in machine learning, separated by the interpolation threshold, in various contexts, has been a central aspect of the discussion in this paper. A concise summary of some of these differences in a single table is given below.
}

\vskip.2in
{\small

\begin{adjustwidth}{-.7cm}{}
\begin{tabular}{l|c|c}
\multicolumn{1}{c}{}&\multicolumn{2}{|l}{
~~\,\includegraphics[height=1.6in,width=4.59in]{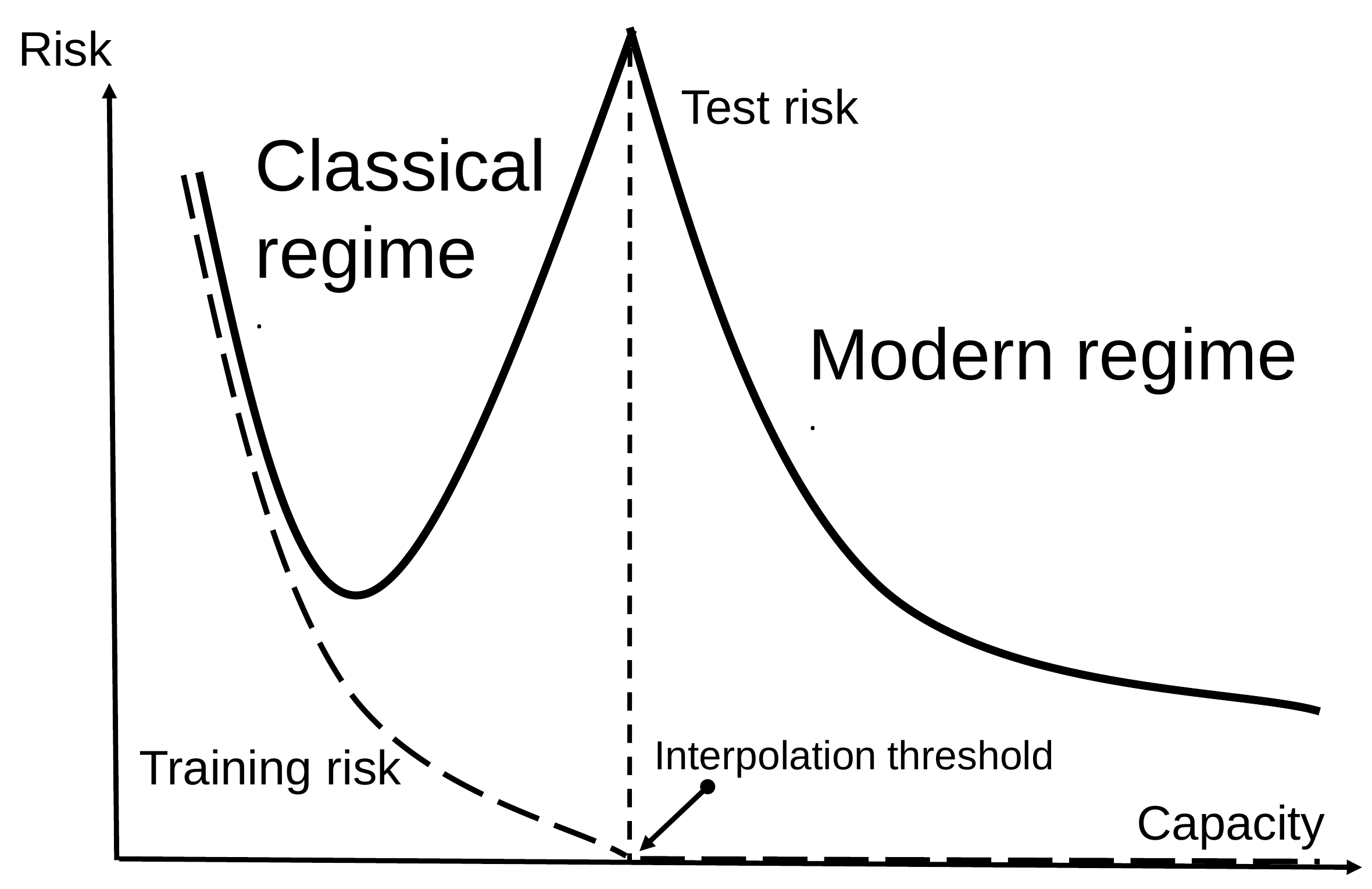}
} \\
&Classical  (under-parameterized)&Modern (over-parameterized)\\
\hline
Generalization curve&U-shaped & Descending  \\
\hline
Optimal model & Bottom of U (hard to find)& Any large model (easy to find)\\ 
\hline
Optimization landscape:& Locally convex & Not locally convex\\
&Minimizers locally unique & Manifolds of minimizers \\ 
&&Satisfies PL* condition\\
\hline
GD/SGD convergence&GD converges to local min&GD/SGD converge to global min\\ 
&SGD w. fixed learning rate does& SGD w. fixed learning rate \\
&not converge&converges exponentially\\
\hline
Adversarial examples&?&Unavoidable \\
\hline
Transition to linearity&& Wide networks w. linear last layer\\
\end{tabular}
 \end{adjustwidth}
}


\subsection{Through a glass darkly}\label{sec:through}

In conclusion, it may be worthwhile to discuss some of the many missing or nebulous mathematical pieces in the gradually coalescing  jigsaw puzzle of deep learning.  


\paragraph{Inverse and direct methods.}
To my mind, the most puzzling question of machine learning is why inverse methods, requiring optimization or inversion, generally perform better than direct methods such as nearest neighbors. 
For example, a kernel machine with a positive definite kernel $K(\bx,\bz)$, appears to perform consistently and measurably better than a Nadaraya-Watson (NW) classifier using the same kernel (or the same family of kernels), despite the fact that both have the same functional form 
$$
f(\bx)= \sum_{i=1}^n \alpha_i K(\bx_i,\bx),\, \alpha_i \in \R
$$
The difference is that for a kernel machine $\balpha=(K)^{-1}\by$, which requires a kernel matrix inversion\footnote{Regularization, e.g., $\balpha=(K+\lambda I)^{-1}\by$ does not change the nature of the problem.}, while NW \brown{(for classification)} simply puts $\balpha =\by$.

The advantage of inverse methods appears to be a broad empirical pattern,  manifested, in particular, by successes of neural networks. Indeed, were it not the case that inverse methods performed significantly better, the Machine Learning landscape would look quite different -- there would be far less need for optimization techniques and, likely, less dependence on the availability of computational resources.   
I am not aware of any compelling theoretical analyses to explain this remarkable empirical difference.

\paragraph{Why does optimization align with statistics?} 
A related question is that of the inductive bias. In over-parameterized settings,  optimization methods, such as commonly used SGD and Adam~\cite{kingma2014adam}, 
select  a  specific point $\bw^*$ in the set of  parameters $\S$ corresponding to interpolating solutions. In fact,  given that $\bw^*$ depends on the initialization typically chosen randomly, e.g., from a normal distribution, 
we should view $\bw^*$ as sampled from some induced probability distribution $\mu$ on the subset of $\S$ reachable by optimization.

Why do parameters sampled from $\mu$ consistently generalize to  data unseen in training?

While there is significant recent work on this topic,  including a number of papers cited  here, and the picture is becoming clearer for linear and kernel methods, we are still far from a thorough theoretical understanding of this alignment in  general deep learning.
Note that interpolation is particularly helpful in addressing this question as it removes the extra complication of analyzing the trade-off between the inductive bias and the empirical loss.  

\paragraph{Early stopping vs. interpolation.} In this paper we have concentrated on interpolation as it provides insights into the phenomena of deep learning. Yet, in practice, at least for neural networks, precise interpolation is rarely used. Instead, iterative optimization algorithms are typically stopped when the validation loss stops decreasing or according to some other {\it early stopping} criterion. 

This is done both for computational reasons, as  running SGD-type algorithms to numerical convergence is typically impractical and unnecessary, but also to improve generalization, as early stopping can be viewed as a type of regularization (e.g.,~\cite{yao2007early}) \blue{or label denoising~\cite{li2020gradient}} that can improve test performance.

For kernel machines with standard Laplacian and Gaussian kernels, a setting where both early stopping and exact solutions can be readily computed,  early stopping seems to provide at best a modest improvement to  generalization performance~\cite{belkin2018understand}.  
Yet, even for kernel machines, computational efficiency of training on larger datasets seems to require iterative methods similar to SGD~\cite{ma2018kernel, meanti2020kernel}, thus making early stopping a computational necessity. 

Despite extensive experimental work, the computational and statistical trade-offs of early stopping in the non-convex over-parameterized regimes remain murky.

\paragraph{Are deep neural networks kernel machines?} A remarkable recent theoretical deep learning discovery  (discussed in Section~\ref{sec:ntk}) is that in certain regimes very wide neural networks are equivalent to kernel machines. At this point much of the theoretical discussion centers on understanding the ``rich regimes'' (e.g.,~\cite{woodworth2020kernel,ghorbani2020neural})), often identified with ``feature learning'', \blue{i.e., learning representations from data}. In these regimes,  tangent kernels change during the training, hence neural networks are not approximated by kernel machines, i.e., a feature map followed by a linear method. The prevalent view among both theoreticians and practitioners, is that success of neural networks cannot be explained by kernel methods as kernel predictors. Yet kernel change during training does not logically imply useful learning and may be an extraneous side effect. \brown{Thus the the question of equivalence remains open. }
Recent, more
sophisticated, kernel machines show performance much closer to the state-of-the-art on certain tasks~\cite{arora2019harnessing, shankar2020neural} but have not yet closed the gap with neural networks.

Without going into a detailed analysis of the arguments (unlikely to be fruitful in any case, as performance of networks has not been conclusively matched by kernels, \brown{nor} is there a convincing ``smoking gun'' argument why it cannot be), it is worth outlining three possibilities:
\begin{itemize}
    \item Neural network performance has elements which cannot \brown{be} replicated by kernel machines (linear optimization problems). 
    \item Neural networks can be approximated by data-dependent kernels, where the kernel function and the Reproducing Kernel Hilbert Space depend on the training data
    (e.g., \brown{on} unlabeled training data like ``warped RKHS''~\cite{sindhwani2005beyond}). 
    \item Neural networks in practical settings can be effectively approximated by kernels, such as Neural Tangent Kernels corresponding to infinitely wide networks~\cite{jacot2018neural}.
\end{itemize}

\blue{I am hopeful that in the near future some \orange{clarity} on these points will be achieved.}

\paragraph{\blue{The role of depth.}}  
\blue{Last and, possibly, least,  we would be remiss to  ignore the question of depth in a paper with {\it deep} in its title. Yet, while many analyses in this paper are applicable to multi-layered networks, it is the width that drives most of the observed phenomena and intuitions. Despite recent efforts, the importance of depth is still not well-understood. Properties of deep architectures  point to the limitations of simple parameter counting -- increasing the depth of an architecture appears to have very different effects from increasing the width, even if the total number of  trainable parameters is the same. In particular, while wider networks are generally observed to perform better than more narrow architectures (\brown{\cite{lee2020finite}}, even with optimal early stopping~\cite{nakkiran2019deep}), the same is not true with respect to the depth, where very deep architectures \orange{can be} inferior~\cite{nichani2020deeper}. One line of inquiry is interpreting depth recursively. Indeed,
in certain settings increasing the depth manifests similarly to iterating a map given by a shallow network~\cite{radha2020over}. Furthermore, fixed points of such iterations have been proposed as an alternative to deep networks with some success~\cite{bai2019deep}. 
More weight for this point of view is provided by the fact that tangent kernels of infinitely wide architectures satisfy a recursive relationship with respect to their depth~\cite{jacot2018neural}.
}

\green{
\section*{Acknowledgements}
A version of this work will appear in Acta Numerica. I would like to thank Acta Numerica for the invitation to write this article and its careful editing.  I thank  Daniel Hsu, Chaoyue Liu, Adityanarayanan Radhakrishnan, Steven Wright and Libin Zhu  for reading the draft and providing numerous helpful suggestions and corrections. I am especially grateful to Daniel Hsu and Steven Wright  for insightful comments which helped clarify exposition of key concepts.
}
\green{
The perspective outlined here has been influenced and informed by many illuminating discussions with  collaborators, colleagues, and students. Many of these discussions  occurred in spring 2017 and summer 2019 during two excellent programs  on foundations of deep learning  at the Simons Institute for the Theory of Computing at Berkeley. I thank it for the hospitality. 
}
\green{Finally, I thank the National Science Foundation and the Simons Foundation for financial support.
}

\bibliographystyle{plain}
\bibliography{main}

\begin{thebibliography}{100}

\bibitem{allen2020backward}
Zeyuan Allen-Zhu and Yuanzhi Li.
\newblock Backward feature correction: How deep learning performs deep
  learning.
\newblock {\em arXiv preprint arXiv:2001.04413}, 2020.

\bibitem{arora2019harnessing}
Sanjeev Arora, Simon~S. Du, Zhiyuan Li, Ruslan Salakhutdinov, Ruosong Wang, and
  Dingli Yu.
\newblock Harnessing the power of infinitely wide deep nets on small-data
  tasks.
\newblock In {\em International Conference on Learning Representations}, 2020.

\bibitem{bai2019deep}
Shaojie Bai, J~Zico Kolter, and Vladlen Koltun.
\newblock Deep equilibrium models.
\newblock {\em Advances in Neural Information Processing Systems}, 32:690--701,
  2019.

\bibitem{bartlett2020failures}
Peter~L Bartlett and Philip~M Long.
\newblock Failures of model-dependent generalization bounds for least-norm
  interpolation.
\newblock {\em arXiv preprint arXiv:2010.08479}, 2020.

\bibitem{bartlett2020benign}
Peter~L Bartlett, Philip~M Long, G{\'a}bor Lugosi, and Alexander Tsigler.
\newblock Benign overfitting in linear regression.
\newblock {\em Proceedings of the National Academy of Sciences}, 2020.

\bibitem{bartlett2002rademacher}
Peter~L Bartlett and Shahar Mendelson.
\newblock Rademacher and gaussian complexities: Risk bounds and structural
  results.
\newblock {\em Journal of Machine Learning Research}, 3(Nov):463--482, 2002.

\bibitem{bartlett2021deep}
Peter~L. Bartlett, Andrea Montanari, and Alexander Rakhlin.
\newblock Deep learning: a statistical viewpoint, 2021.

\bibitem{bassily2018exponential}
Raef Bassily, Mikhail Belkin, and Siyuan Ma.
\newblock On exponential convergence of sgd in non-convex over-parametrized
  learning.
\newblock {\em arXiv preprint arXiv:1811.02564}, 2018.

\bibitem{belkin2019reconciling}
Mikhail Belkin, Daniel Hsu, Siyuan Ma, and Soumik Mandal.
\newblock Reconciling modern machine-learning practice and the classical
  bias{\textendash}variance trade-off.
\newblock {\em Proceedings of the National Academy of Sciences},
  116(32):15849--15854, 2019.

\bibitem{belkin2018overfitting}
Mikhail Belkin, Daniel Hsu, and Partha Mitra.
\newblock Overfitting or perfect fitting? risk bounds for classification and
  regression rules that interpolate.
\newblock In {\em Advances in Neural Information Processing Systems}, pages
  2306--2317, 2018.

\bibitem{belkin2020two}
Mikhail Belkin, Daniel Hsu, and Ji~Xu.
\newblock Two models of double descent for weak features.
\newblock {\em SIAM Journal on Mathematics of Data Science}, 2(4):1167--1180,
  2020.

\bibitem{belkin2018understand}
Mikhail Belkin, Siyuan Ma, and Soumik Mandal.
\newblock To understand deep learning we need to understand kernel learning.
\newblock In {\em Proceedings of the 35th International Conference on Machine
  Learning}, volume~80 of {\em Proceedings of Machine Learning Research}, pages
  541--549, 2018.

\bibitem{belkin2018does}
Mikhail Belkin, Alexander Rakhlin, and Alexandre~B Tsybakov.
\newblock https://arxiv.org/abs/1806.09471, 2018.

\bibitem{blumer1987occam}
Anselm Blumer, Andrzej Ehrenfeucht, David Haussler, and Manfred~K Warmuth.
\newblock Occam's razor.
\newblock {\em Information processing letters}, 24(6):377--380, 1987.

\bibitem{bousquet2003introduction}
Olivier Bousquet, St{\'e}phane Boucheron, and G{\'a}bor Lugosi.
\newblock Introduction to statistical learning theory.
\newblock In {\em Summer School on Machine Learning}, pages 169--207. Springer,
  2003.

\bibitem{breiman1995reflections}
Leo Breiman.
\newblock Reflections after refereeing papers for nips.
\newblock {\em The Mathematics of Generalization}, pages 11--15, 1995.

\bibitem{bubeck2015convex}
S{\'e}bastien Bubeck et~al.
\newblock Convex optimization: Algorithms and complexity.
\newblock {\em Foundations and Trends{\textregistered} in Machine Learning},
  8(3-4):231--357, 2015.

\bibitem{buja2007comment}
Andreas Buja, David Mease, Abraham~J Wyner, et~al.
\newblock Comment: Boosting algorithms: Regularization, prediction and model
  fitting.
\newblock {\em Statistical Science}, 22(4):506--512, 2007.

\bibitem{canziani2016analysis}
Alfredo Canziani, Adam Paszke, and Eugenio Culurciello.
\newblock An analysis of deep neural network models for practical applications.
\newblock {\em arXiv preprint arXiv:1605.07678}, 2016.

\bibitem{cover1967nearest}
Thomas Cover and Peter Hart.
\newblock Nearest neighbor pattern classification.
\newblock {\em IEEE transactions on information theory}, 13(1):21--27, 1967.

\bibitem{cutler2001pert}
Adele Cutler and Guohua Zhao.
\newblock Pert-perfect random tree ensembles.
\newblock {\em Computing Science and Statistics}, 33:490--497, 2001.

\bibitem{defazio2014saga}
Aaron Defazio, Francis Bach, and Simon Lacoste-Julien.
\newblock Saga: A fast incremental gradient method with support for
  non-strongly convex composite objectives.
\newblock In {\em NIPS}, pages 1646--1654, 2014.

\bibitem{devroye1998hilbert}
Luc Devroye, Laszlo Gy{\"o}rfi, and Adam Krzy{\.z}ak.
\newblock The hilbert kernel regression estimate.
\newblock {\em Journal of Multivariate Analysis}, 65(2):209--227, 1998.

\bibitem{du2018gradientdeep}
Simon Du, Jason Lee, Haochuan Li, Liwei Wang, and Xiyu Zhai.
\newblock Gradient descent finds global minima of deep neural networks.
\newblock In {\em International Conference on Machine Learning}, pages
  1675--1685, 2019.

\bibitem{du2018gradientshallow}
Simon~S Du, Xiyu Zhai, Barnabas Poczos, and Aarti Singh.
\newblock Gradient descent provably optimizes over-parameterized neural
  networks.
\newblock {\em arXiv preprint arXiv:1810.02054}, 2018.

\bibitem{fawzi2016robustness}
Alhussein Fawzi, Seyed-Mohsen Moosavi-Dezfooli, and Pascal Frossard.
\newblock Robustness of classifiers: from adversarial to random noise.
\newblock In {\em Advances in Neural Information Processing Systems}, pages
  1632--1640, 2016.

\bibitem{fedus2021switch}
William Fedus, Barret Zoph, and Noam Shazeer.
\newblock Switch transformers: Scaling to trillion parameter models with simple
  and efficient sparsity, 2021.

\bibitem{freund1997decision}
Yoav Freund and Robert~E Schapire.
\newblock A decision-theoretic generalization of on-line learning and an
  application to boosting.
\newblock {\em Journal of computer and system sciences}, 55(1):119--139, 1997.

\bibitem{geman1992neural}
Stuart Geman, Elie Bienenstock, and René Doursat.
\newblock Neural networks and the bias/variance dilemma.
\newblock {\em Neural Computation}, 4(1):1--58, 1992.

\bibitem{ghorbani2020neural}
Behrooz Ghorbani, Song Mei, Theodor Misiakiewicz, and Andrea Montanari.
\newblock When do neural networks outperform kernel methods?
\newblock In Hugo Larochelle, Marc'Aurelio Ranzato, Raia Hadsell,
  Maria{-}Florina Balcan, and Hsuan{-}Tien Lin, editors, {\em Advances in
  Neural Information Processing Systems 33: Annual Conference on Neural
  Information Processing Systems 2020, NeurIPS 2020, December 6-12, 2020,
  virtual}, 2020.

\bibitem{Goodfellow-et-al-2016}
Ian Goodfellow, Yoshua Bengio, and Aaron Courville.
\newblock {\em Deep Learning}.
\newblock MIT Press, 2016.

\bibitem{gyorfi02}
L{\'{a}}szl{\'{o}} Gy{\"{o}}rfi, Michael Kohler, Adam Krzyzak, and Harro Walk.
\newblock {\em A Distribution-Free Theory of Nonparametric Regression}.
\newblock Springer series in statistics. Springer, 2002.

\bibitem{halton1991simplicial}
John~H Halton.
\newblock Simplicial multivariable linear interpolation.
\newblock Technical Report TR91-002, University of North Carolina at Chapel
  Hill, Department of Computer Science, 1991.

\bibitem{hastie2019surprises}
Trevor Hastie, Andrea Montanari, Saharon Rosset, and Ryan~J Tibshirani.
\newblock Surprises in high-dimensional ridgeless least squares interpolation.
\newblock {\em arXiv preprint arXiv:1903.08560}, 2019.

\bibitem{friedman2001elements}
Trevor Hastie, Robert Tibshirani, and Jerome Friedman.
\newblock {\em The Elements of Statistical Learning}, volume~1.
\newblock Springer, 2001.

\bibitem{hui2021evaluation}
Like Hui and Mikhail Belkin.
\newblock Evaluation of neural architectures trained with square loss vs
  cross-entropy in classification tasks.
\newblock In {\em International Conference on Learning Representations}, 2021.

\bibitem{ilyas2019adversarial}
Andrew Ilyas, Shibani Santurkar, Logan Engstrom, Brandon Tran, and Aleksander
  Madry.
\newblock Adversarial examples are not bugs, they are features.
\newblock {\em Advances in neural information processing systems}, 32, 2019.

\bibitem{jacot2018neural}
Arthur Jacot, Franck Gabriel, and Cl{\'e}ment Hongler.
\newblock Neural tangent kernel: Convergence and generalization in neural
  networks.
\newblock In {\em Advances in neural information processing systems}, pages
  8571--8580, 2018.

\bibitem{ji2019implicit}
Ziwei Ji and Matus Telgarsky.
\newblock The implicit bias of gradient descent on nonseparable data.
\newblock In Alina Beygelzimer and Daniel Hsu, editors, {\em Proceedings of the
  Thirty-Second Conference on Learning Theory}, volume~99 of {\em Proceedings
  of Machine Learning Research}, pages 1772--1798, Phoenix, USA, 25--28 Jun
  2019. PMLR.

\bibitem{johnson2013accelerating}
Rie Johnson and Tong Zhang.
\newblock Accelerating stochastic gradient descent using predictive variance
  reduction.
\newblock In {\em NIPS}, pages 315--323, 2013.

\bibitem{kaczmarz1937angenaherte}
Stefan Kaczmarz.
\newblock Angenaherte auflosung von systemen linearer gleichungen.
\newblock {\em Bull. Int. Acad. Sci. Pologne, A}, 35, 1937.

\bibitem{karimi2016linear}
Hamed Karimi, Julie Nutini, and Mark Schmidt.
\newblock Linear convergence of gradient and proximal-gradient methods under
  the polyak-lojasiewicz condition.
\newblock In {\em Joint European Conference on Machine Learning and Knowledge
  Discovery in Databases}, pages 795--811. Springer, 2016.

\bibitem{kimeldorf1970correspondence}
George~S Kimeldorf and Grace Wahba.
\newblock A correspondence between bayesian estimation on stochastic processes
  and smoothing by splines.
\newblock {\em The Annals of Mathematical Statistics}, 41(2):495--502, 1970.

\bibitem{kingma2014adam}
Diederik~P. Kingma and Jimmy Ba.
\newblock Adam: {A} method for stochastic optimization.
\newblock In Yoshua Bengio and Yann LeCun, editors, {\em 3rd International
  Conference on Learning Representations, {ICLR} 2015, San Diego, CA, USA, May
  7-9, 2015, Conference Track Proceedings}, 2015.

\bibitem{streetlight}
Yann Lecun.
\newblock The epistemology of deep learning.
\newblock \url{https://www.youtube.com/watch?v=gG5NCkMerHU&t=3210s}.

\bibitem{lee2020finite}
Jaehoon Lee, Samuel~S Schoenholz, Jeffrey Pennington, Ben Adlam, Lechao Xiao,
  Roman Novak, and Jascha Sohl-Dickstein.
\newblock Finite versus infinite neural networks: an empirical study.
\newblock {\em arXiv preprint arXiv:2007.15801}, 2020.

\bibitem{lee2019wide}
Jaehoon Lee, Lechao Xiao, Samuel Schoenholz, Yasaman Bahri, Roman Novak, Jascha
  Sohl-Dickstein, and Jeffrey Pennington.
\newblock Wide neural networks of any depth evolve as linear models under
  gradient descent.
\newblock In {\em Advances in neural information processing systems}, pages
  8570--8581, 2019.

\bibitem{li2020gradient}
Mingchen Li, Mahdi Soltanolkotabi, and Samet Oymak.
\newblock Gradient descent with early stopping is provably robust to label
  noise for overparameterized neural networks.
\newblock In {\em International Conference on Artificial Intelligence and
  Statistics}, pages 4313--4324. PMLR, 2020.

\bibitem{liang2020just}
Tengyuan Liang, Alexander Rakhlin, et~al.
\newblock Just interpolate: Kernel ridgeless regression can generalize.
\newblock {\em Annals of Statistics}, 48(3):1329--1347, 2020.

\bibitem{liu2018accelerating}
Chaoyue Liu and Mikhail Belkin.
\newblock Accelerating sgd with momentum for over-parameterized learning.
\newblock In {\em The 8th International Conference on Learning Representations
  (ICLR)}, 2020.

\bibitem{liu2020toward}
Chaoyue Liu, Libin Zhu, and Mikhail Belkin.
\newblock Loss landscapes and optimization in over-parameterized non-linear
  systems and neural networks.
\newblock {\em arXiv preprint arXiv:2003.00307}, 2020.

\bibitem{liu2020linearity}
Chaoyue Liu, Libin Zhu, and Mikhail Belkin.
\newblock On the linearity of large non-linear models: when and why the tangent
  kernel is constant.
\newblock {\em Advances in Neural Information Processing Systems}, 33, 2020.

\bibitem{lojasiewicz1963topological}
Stanislaw Lojasiewicz.
\newblock A topological property of real analytic subsets.
\newblock {\em Coll. du CNRS, Les {\'e}quations aux d{\'e}riv{\'e}es
  partielles}, 117:87--89, 1963.

\bibitem{loog2020brief}
Marco Loog, Tom Viering, Alexander Mey, Jesse~H Krijthe, and David~MJ Tax.
\newblock A brief prehistory of double descent.
\newblock {\em Proceedings of the National Academy of Sciences},
  117(20):10625--10626, 2020.

\bibitem{ma2018power}
Siyuan Ma, Raef Bassily, and Mikhail Belkin.
\newblock The power of interpolation: Understanding the effectiveness of {SGD}
  in modern over-parametrized learning.
\newblock In {\em Proceedings of the 35th International Conference on Machine
  Learning, {ICML} 2018, Stockholmsm{\"{a}}ssan, Stockholm, Sweden, July 10-15,
  2018}, volume~80 of {\em Proceedings of Machine Learning Research}, pages
  3331--3340. {PMLR}, 2018.

\bibitem{ma2018kernel}
Siyuan Ma and Mikhail Belkin.
\newblock Kernel machines that adapt to gpus for effective large batch
  training.
\newblock In A.~Talwalkar, V.~Smith, and M.~Zaharia, editors, {\em Proceedings
  of Machine Learning and Systems}, volume~1, pages 360--373, 2019.

\bibitem{madry2017towards}
Aleksander Madry, Aleksandar Makelov, Ludwig Schmidt, Dimitris Tsipras, and
  Adrian Vladu.
\newblock Towards deep learning models resistant to adversarial attacks.
\newblock In {\em International Conference on Learning Representations}, 2018.

\bibitem{mai2019high}
Xiaoyi Mai and Zhenyu Liao.
\newblock High dimensional classification via empirical risk minimization:
  Improvements and optimality.
\newblock {\em arXiv preprint arXiv:1905.13742}, 2019.

\bibitem{meanti2020kernel}
Giacomo Meanti, Luigi Carratino, Lorenzo Rosasco, and Alessandro Rudi.
\newblock Kernel methods through the roof: handling billions of points
  efficiently.
\newblock {\em arXiv preprint arXiv:2006.10350}, 2020.

\bibitem{mei2019generalization}
Song Mei and Andrea Montanari.
\newblock The generalization error of random features regression: Precise
  asymptotics and double descent curve.
\newblock {\em arXiv preprint arXiv:1908.05355}, 2019.

\bibitem{mitra2019understanding}
Partha~P Mitra.
\newblock Understanding overfitting peaks in generalization error: Analytical
  risk curves for $\ell_2 $ and $\ell_1 $ penalized interpolation.
\newblock {\em arXiv preprint arXiv:1906.03667}, 2019.

\bibitem{moulines2011non}
Eric Moulines and Francis~R Bach.
\newblock Non-asymptotic analysis of stochastic approximation algorithms for
  machine learning.
\newblock In {\em Advances in Neural Information Processing Systems}, pages
  451--459, 2011.

\bibitem{muthukumar2020classification}
Vidya Muthukumar, Adhyyan Narang, Vignesh Subramanian, Mikhail Belkin, Daniel
  Hsu, and Anant Sahai.
\newblock Classification vs regression in overparameterized regimes: Does the
  loss function matter?, 2020.

\bibitem{muthukumar2020harmless}
Vidya Muthukumar, Kailas Vodrahalli, Vignesh Subramanian, and Anant Sahai.
\newblock Harmless interpolation of noisy data in regression.
\newblock {\em IEEE Journal on Selected Areas in Information Theory}, 2020.

\bibitem{nadaraya1964estimating}
Elizbar~A Nadaraya.
\newblock On estimating regression.
\newblock {\em Theory of Probability \& Its Applications}, 9(1):141--142, 1964.

\bibitem{nagarajan2019uniform}
Vaishnavh Nagarajan and J.~Zico Kolter.
\newblock Uniform convergence may be unable to explain generalization in deep
  learning.
\newblock In {\em Advances in Neural Information Processing Systems},
  volume~32, 2019.

\bibitem{nakkiran2019deep}
Preetum Nakkiran, Gal Kaplun, Yamini Bansal, Tristan Yang, Boaz Barak, and Ilya
  Sutskever.
\newblock Deep double descent: Where bigger models and more data hurt.
\newblock In {\em International Conference on Learning Representations}, 2019.

\bibitem{needell2014stochastic}
Deanna Needell, Rachel Ward, and Nati Srebro.
\newblock Stochastic gradient descent, weighted sampling, and the randomized
  kaczmarz algorithm.
\newblock In {\em NIPS}, 2014.

\bibitem{negrea2020indefense}
Jeffrey Negrea, Gintare~Karolina Dziugaite, and Daniel Roy.
\newblock In defense of uniform convergence: Generalization via derandomization
  with an application to interpolating predictors.
\newblock In Hal~Daumé III and Aarti Singh, editors, {\em Proceedings of the
  37th International Conference on Machine Learning}, volume 119 of {\em
  Proceedings of Machine Learning Research}, pages 7263--7272. PMLR, 13--18 Jul
  2020.

\bibitem{neyshabur2015search}
Behnam Neyshabur, Ryota Tomioka, and Nathan Srebro.
\newblock In search of the real inductive bias: On the role of implicit
  regularization in deep learning.
\newblock In {\em ICLR (Workshop)}, 2015.

\bibitem{nichani2020deeper}
Eshaan Nichani, Adityanarayanan Radhakrishnan, and Caroline Uhler.
\newblock Do deeper convolutional networks perform better?
\newblock {\em arXiv preprint arXiv:2010.09610}, 2020.

\bibitem{nocedal2006numerical}
Jorge Nocedal and Stephen Wright.
\newblock {\em Numerical optimization}.
\newblock Springer Science \& Business Media, 2006.

\bibitem{oymak2019overparameterized}
Samet Oymak and Mahdi Soltanolkotabi.
\newblock Overparameterized nonlinear learning: Gradient descent takes the
  shortest path?
\newblock In {\em International Conference on Machine Learning}, pages
  4951--4960. PMLR, 2019.

\bibitem{polyak1963gradient}
Boris~Teodorovich Polyak.
\newblock Gradient methods for minimizing functionals.
\newblock {\em Zhurnal Vychislitel'noi Matematiki i Matematicheskoi Fiziki},
  3(4):643--653, 1963.

\bibitem{pravesh2020expressive}
Kothari~K Pravesh and Livni Roi.
\newblock On the expressive power of kernel methods and the efficiency of
  kernel learning by association schemes.
\newblock In {\em Algorithmic Learning Theory}, pages 422--450. PMLR, 2020.

\bibitem{radha2020over}
Adityanarayanan Radhakrishnan, Mikhail Belkin, and Caroline Uhler.
\newblock Overparameterized neural networks implement associative memory.
\newblock {\em Proceedings of the National Academy of Sciences},
  117(44):27162--27170, 2020.

\bibitem{alchemy}
Ali Rahimi and Ben. Recht.
\newblock Reflections on random kitchen sinks.
\newblock \url{http://www.argmin.net/2017/12/05/kitchen-sinks/}, 2017.

\bibitem{rahimi2008random}
Ali Rahimi and Benjamin Recht.
\newblock Random features for large-scale kernel machines.
\newblock In {\em Advances in Neural Information Processing Systems}, pages
  1177--1184, 2008.

\bibitem{rifkin2002everything}
Ryan~Michael Rifkin.
\newblock {\em Everything old is new again: a fresh look at historical
  approaches in machine learning}.
\newblock PhD thesis, Massachusetts Institute of Technology, 2002.

\bibitem{roux2012stochastic}
Nicolas~L Roux, Mark Schmidt, and Francis~R Bach.
\newblock A stochastic gradient method with an exponential convergence rate for
  finite training sets.
\newblock In {\em NIPS}, pages 2663--2671, 2012.

\bibitem{ruslantutorial}
Ruslan Salakhutdinov.
\newblock Tutorial on deep learning.
\newblock
  \url{https://simons.berkeley.edu/talks/ruslan-salakhutdinov-01-26-2017-1}.

\bibitem{schapire1998}
Robert~E. Schapire, Yoav Freund, Peter Bartlett, and Wee~Sun Lee.
\newblock Boosting the margin: a new explanation for the effectiveness of
  voting methods.
\newblock {\em Ann. Statist.}, 26(5):1651--1686, 1998.

\bibitem{senior2020improved}
Andrew Senior, Richard Evans, John Jumper, James Kirkpatrick, Laurent Sifre,
  Tim Green, Chongli Qin, Augustin Zidek, Alexander~WR Nelson, Alex Bridgland,
  et~al.
\newblock Improved protein structure prediction using potentials from deep
  learning.
\newblock {\em Nature}, 577(7792):706--710, 2020.

\bibitem{shankar2020neural}
Vaishaal Shankar, Alex Fang, Wenshuo Guo, Sara Fridovich-Keil, Jonathan
  Ragan-Kelley, Ludwig Schmidt, and Benjamin Recht.
\newblock Neural kernels without tangents.
\newblock In {\em Proceedings of the 37th International Conference on Machine
  Learning}, volume 119, pages 8614--8623. PMLR, 2020.

\bibitem{shepard1968two}
Donald Shepard.
\newblock A two-dimensional interpolation function for irregularly-spaced data.
\newblock In {\em Proceedings of the 1968 23rd ACM national conference}, pages
  517--524, 1968.

\bibitem{sindhwani2005beyond}
Vikas Sindhwani, Partha Niyogi, and Mikhail Belkin.
\newblock Beyond the point cloud: from transductive to semi-supervised
  learning.
\newblock In {\em Proceedings of the 22nd international conference on Machine
  learning}, pages 824--831, 2005.

\bibitem{spigler2018jamming}
S~Spigler, M~Geiger, S~d'Ascoli, L~Sagun, G~Biroli, and M~Wyart.
\newblock A jamming transition from under- to over-parametrization affects
  generalization in deep learning.
\newblock {\em Journal of Physics A: Mathematical and Theoretical},
  52(47):474001, oct 2019.

\bibitem{Gauss_least_squares}
Stephen~M. Stigler.
\newblock {Gauss and the Invention of Least Squares}.
\newblock {\em The Annals of Statistics}, 9(3):465 -- 474, 1981.

\bibitem{strohmer2009randomized}
Thomas Strohmer and Roman Vershynin.
\newblock A randomized kaczmarz algorithm with exponential convergence.
\newblock {\em Journal of Fourier Analysis and Applications}, 15(2), 2009.

\bibitem{su2017one}
Jiawei Su, Danilo~Vasconcellos Vargas, and Kouichi Sakurai.
\newblock One pixel attack for fooling deep neural networks.
\newblock {\em IEEE Transactions on Evolutionary Computation}, 23(5):828--841,
  2019.

\bibitem{szegedy2013adversarial}
Christian Szegedy, Wojciech Zaremba, Ilya Sutskever, Joan Bruna, Dumitru Erhan,
  Ian Goodfellow, and Rob Fergus.
\newblock Intriguing properties of neural networks.
\newblock In {\em International Conference on Learning Representations}, 2014.

\bibitem{thrampoulidis2020theoretical}
Christos Thrampoulidis, Samet Oymak, and Mahdi Soltanolkotabi.
\newblock Theoretical insights into multiclass classification: A
  high-dimensional asymptotic view.
\newblock In H.~Larochelle, M.~Ranzato, R.~Hadsell, M.~F. Balcan, and H.~Lin,
  editors, {\em Advances in Neural Information Processing Systems}, volume~33,
  pages 8907--8920. Curran Associates, Inc., 2020.

\bibitem{Vapnik}
Vladimir~N. Vapnik.
\newblock {\em The Nature of Statistical Learning Theory}.
\newblock Springer, 1995.

\bibitem{warmuth2005leaving}
Manfred~K Warmuth and SVN Vishwanathan.
\newblock Leaving the span.
\newblock In {\em International Conference on Computational Learning Theory},
  pages 366--381. Springer, 2005.

\bibitem{watson1964smooth}
Geoffrey~S Watson.
\newblock Smooth regression analysis.
\newblock {\em Sankhy{\=a}: The Indian Journal of Statistics, Series A}, pages
  359--372, 1964.

\bibitem{wendland_2004}
Holger Wendland.
\newblock {\em Scattered Data Approximation}.
\newblock Cambridge Monographs on Applied and Computational Mathematics.
  Cambridge University Press, 2004.

\bibitem{woodworth2020kernel}
Blake Woodworth, Suriya Gunasekar, Jason~D Lee, Edward Moroshko, Pedro
  Savarese, Itay Golan, Daniel Soudry, and Nathan Srebro.
\newblock Kernel and rich regimes in overparametrized models.
\newblock In {\em Conference on Learning Theory}, pages 3635--3673. PMLR, 2020.

\bibitem{wyner2017explaining}
Abraham~J Wyner, Matthew Olson, Justin Bleich, and David Mease.
\newblock Explaining the success of adaboost and random forests as
  interpolating classifiers.
\newblock {\em Journal of Machine Learning Research}, 18(48):1--33, 2017.

\bibitem{xu2019number}
Ji~Xu and Daniel Hsu.
\newblock On the number of variables to use in principal component regression.
\newblock {\em Advances in neural information processing systems}, 2019.

\bibitem{yao2007early}
Yuan Yao, Lorenzo Rosasco, and Andrea Caponnetto.
\newblock On early stopping in gradient descent learning.
\newblock {\em Constructive Approximation}, 26(2):289--315, 2007.

\bibitem{zhang2016understanding}
Chiyuan Zhang, Samy Bengio, Moritz Hardt, Benjamin Recht, and Oriol Vinyals.
\newblock Understanding deep learning requires rethinking generalization.
\newblock In {\em International Conference on Learning Representations}, 2017.

\bibitem{zhou2020onuniform}
Lijia Zhou, Danica~J Sutherland, and Nati Srebro.
\newblock On uniform convergence and low-norm interpolation learning.
\newblock In H.~Larochelle, M.~Ranzato, R.~Hadsell, M.~F. Balcan, and H.~Lin,
  editors, {\em Advances in Neural Information Processing Systems}, volume~33,
  pages 6867--6877. Curran Associates, Inc., 2020.

\end{thebibliography}

\end{document}